\documentclass{article}
\pdfoutput=1

\usepackage{arxiv}

\usepackage[utf8]{inputenc} 
\usepackage[T1]{fontenc}    
\usepackage[pdfpagelabels]{hyperref}       
\usepackage{url}            
\usepackage{booktabs}       
\usepackage{amsfonts}       
\usepackage{nicefrac}       
\usepackage{microtype}      
\usepackage{graphicx}
\usepackage{doi}

\usepackage{braket}
\usepackage{xcolor}      
\usepackage{adjustbox}
\usepackage{amsmath,amssymb,amsfonts,amsthm}
\usepackage{threeparttable}
\usepackage{multirow}
\usepackage{algorithm}
\usepackage{algorithmic}

\newtheorem{proposition}{\textbf{Proposition}}
\newtheorem{definition}{\textbf{Definition}}
\newtheorem*{proof*}{Proof}

\newcommand{\CheckedBox}{
	\ooalign{$\Box$\cr\hidewidth\raisebox{0.15ex}{\hspace{0.1em}$\checkmark$}\hidewidth\cr}%
}

\title{Approximated Orthogonal Projection Unit: 
	Stabilizing Regression Network Training Using Natural Gradient}

\author{ 
	\href{https://orcid.org/0000-0002-2900-238X}{\includegraphics[scale=0.06]{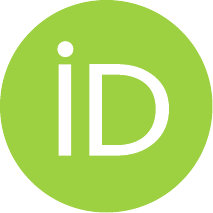}\hspace{1mm}Shaoqi Wang}
	\\
	Control Science and Engineering\\
	Zhejiang University\\
	Hangzhou, 310027 \\
	\texttt{sq\_w@zju.edu.cn} \\
	\And
	\href{https://orcid.org/0000-0002-4362-2104}{\includegraphics[scale=0.06]{orcid.pdf}\hspace{1mm}Chunjie Yang}
	\\
	Control Science and Engineering\\
	Zhejiang University\\
	Hangzhou, 310027 \\
	\texttt{cjyang999@zju.edu.cn} \\
	\And
	\href{https://orcid.org/0000-0001-6611-4754}{\includegraphics[scale=0.06]{orcid.pdf}\hspace{1mm}Siwei Lou}
	\\
	Control Science and Engineering\\
	Zhejiang University\\
	Hangzhou, 310027 \\
	\texttt{swlou@zju.edu.cn} \\
}

\renewcommand{\shorttitle}{\textit{arXiv} Template}

\begin{document}
\maketitle

\begin{abstract}
	Neural networks (NN) are extensively studied in cutting-edge soft sensor models due to their  feature extraction and function approximation capabilities. 
	Current research into network-based methods primarily focuses on models' offline accuracy. 
	Notably, in industrial soft sensor context, online optimizing stability and interpretability are prioritized, followed by accuracy. 
	This requires a clearer understanding of network's training process. 
	To bridge this gap, we propose a novel NN named the Approximated Orthogonal Projection Unit (AOPU) which has solid mathematical basis and presents superior training stability. 
	AOPU truncates the gradient backpropagation at dual parameters, optimizes the trackable parameters updates, and enhances the robustness of training. 
	We further prove that AOPU attains minimum variance estimation (MVE) in NN, wherein the truncated gradient approximates the natural gradient (NG). 
	Empirical results on two chemical process datasets clearly show that AOPU outperforms other models in achieving stable convergence, marking a significant advancement in soft sensor field.
\end{abstract}

\keywords{Neural network \and Stable convergence \and Minimum Variance estimator \and Natural gradient }

\section{Introduction}
Deep learning methods have achieved recent success in many regression areas such as natural language processing, protein structure prediction, and building energy consumption forecasting. However, for these methods to be useful in the industrial soft sensor field, which demands higher immediacy and stability, further research into model structure and the stability of the training process is necessary \cite{RN71,RN56,RN73}. The safety and economic impact of factory impose stringent requirements on soft sensor models deployed online \cite{RN22,RN58}. For example, each mini-batch update must not cause significant performance fluctuations to ensure that downstream controllers and monitors do not execute erroneous actions; soft sensor models must be deployed online to avoid fluctuations due to changes in operating conditions and model switching \cite{OneNet-Enhancing-Time}. Common network's training tricks are not suitable for soft sensor contexts. For example, it's not feasible to use checkpoints for early stopping but to always use the latest updated checkpoint; there wouldn't be adaptive learning rate changes but rather a constant learning rate maintained throughout.
\textcolor{black}{Experimental results demonstrate that such differences lead to a substantial decline in performance.}
These constraints necessitate the development of better-suited network architectures for regression task that ensure more stable optimization during training \cite{RN34,OneNet-Enhancing-Time}.

MVE is the best unbiased estimator under the Mean Squared Error (MSE) criterion, essentially representing the performance ceiling for regression models \cite{On-Multilevel-Best-Linear}. 
Unfortunately, directly applying MVE to NN is challenging due to the difficulty in obtaining the likelihood distribution $p(y|x)$ of inputs $x$ and outputs $y$ \cite{Adaptive-Experimental-Design-with-Temporal-Interference}.
Traditional research on MVE has focused on techniques like Kalman filtering \cite{Stochastic-model-predictive-control,Autodifferentiable-Ensemble-Kalman-Filters}, algorithms based on Ordinary Least Squares (OLS) \cite{RN51,RN69}, and other system identification research \cite{RN38,RN60,RN16} which operate under linear and convex conditions. 
These methods, while effective within their scope, have limited expressive power \cite{RN61}.

Many studies have explored NN optimization from various perspectives such as adaptive learning rates \cite{Decoupled-Weight-Decay-Regularization}, momentum updates \cite{Adam} and customized loss functions \cite{RN64}. 
Compared to the first-order optimization methods, second-order optimization algorithms based on Natural Gradient Descent (NGD) \cite{RN10,RN8} can achieve faster and more stable convergence \cite{RN7}. 
This is because NGD considers the manifold distribution of model parameters by computing the Fisher Information Matrix (FIM) during the gradient update process. 
However, calculating FIM introduces significant computational overhead, making NGD challenging to implement in most machine learning models \cite{Gaussian-Processes-for-Big-Data}. Much research is focused on adapting model structures for NGD \cite{RN45,RN49,Deep-Neural-Networks-as-Point-Estimates} and reducing its computational costs \cite{Fast-and-Simple-Natural-Gradient,Tractable-structured-natural-gradient,Conjugate-Computation-Variational-Inference}, yet applying NGD to NN optimization remains an unsolved issue \cite{Optimizing-Neural-Networks-with-Kronecker-factored,Can-We-Remove,Structured-Inverse-Free}.

Some studies have approached the regression task from the perspective of targeted modular design \cite{RN18,RN28,RN54}, emphasizing the construction of local neuron-level rules to assist models in learning conducive features , exemplified by SLSTM \cite{Nonlinear-Dynamic-Soft-Sensor}, SIAE \cite{Stacked-isomorphic-autoencoder}, MIF-Autoformer \cite{RN20}, and CBMP \cite{RN70}.
Some endeavors have yielded results with solid theoretical underpinnings \cite{RN14,RN58,RN19}, such as S4 \cite{RN29}, VAE, VIOG \cite{RN50}. 
These examples, integrate with other great work \cite{Deep-Neural-Networks-as-Point-Estimates,Synthesizer,RN19,Efficiently-sampling-functions-from-Gaussian}, jointly demonstrate excellent integrations of NN with theoretical basis and also have a stronger expressivity compared to the identification algorithms \cite{RN38,RN60,RN59}. 
However, even though these networks possess certain interpretability, it is still unclear whether these prior biases are beneficial for regression task \cite{Understanding-Recurrent-Neural-Networks-Using-Nonequilibrium-Response-Theory,RN21}.
Furthermore, their mathematical foundations are not rooted in soft sensor tasks, which do not guarantee stable performance in regression \cite{RN13}.

AOPU differs from conventional studies by focusing on training optimization and the overall input-output relationships. 
Assuming there is a robust feature extraction module (augmentation block), AOPU pays specific attention on better optimization and more stable convergence. 
AOPU innovatively introduces trackable and dual parameters, enhancing an structure approximation of MVE. 
The dual parameters are injective representations of trackable parameters, mainly aiding in truncating the gradient backpropagation process.
This truncation will be validated as an effective approximation to NGD.
The augmentation module boosts AOPU's performance and also acts as an extension interface, making AOPU more versatile.
Rank Ratio (RR) is introduced as an interpretability index to provide deep and comprehensive insights into the network dynamics.
RR quantifies the ratio of linearly independent samples within a batch, providing a measure of the heterogeneity and diversity of information. 
By harnessing RR value, we can roughly foresee the performance in advance of the training. 
A high RR suggests the model output more closely approximates the MVE, and optimization is more in line with NGD, leading to superior performance. 
Conversely, a low RR implies that the precision of computation is compromised, resulting in inferior performance.

\begin{figure}
	\centering
	\includegraphics[width=0.6\columnwidth]{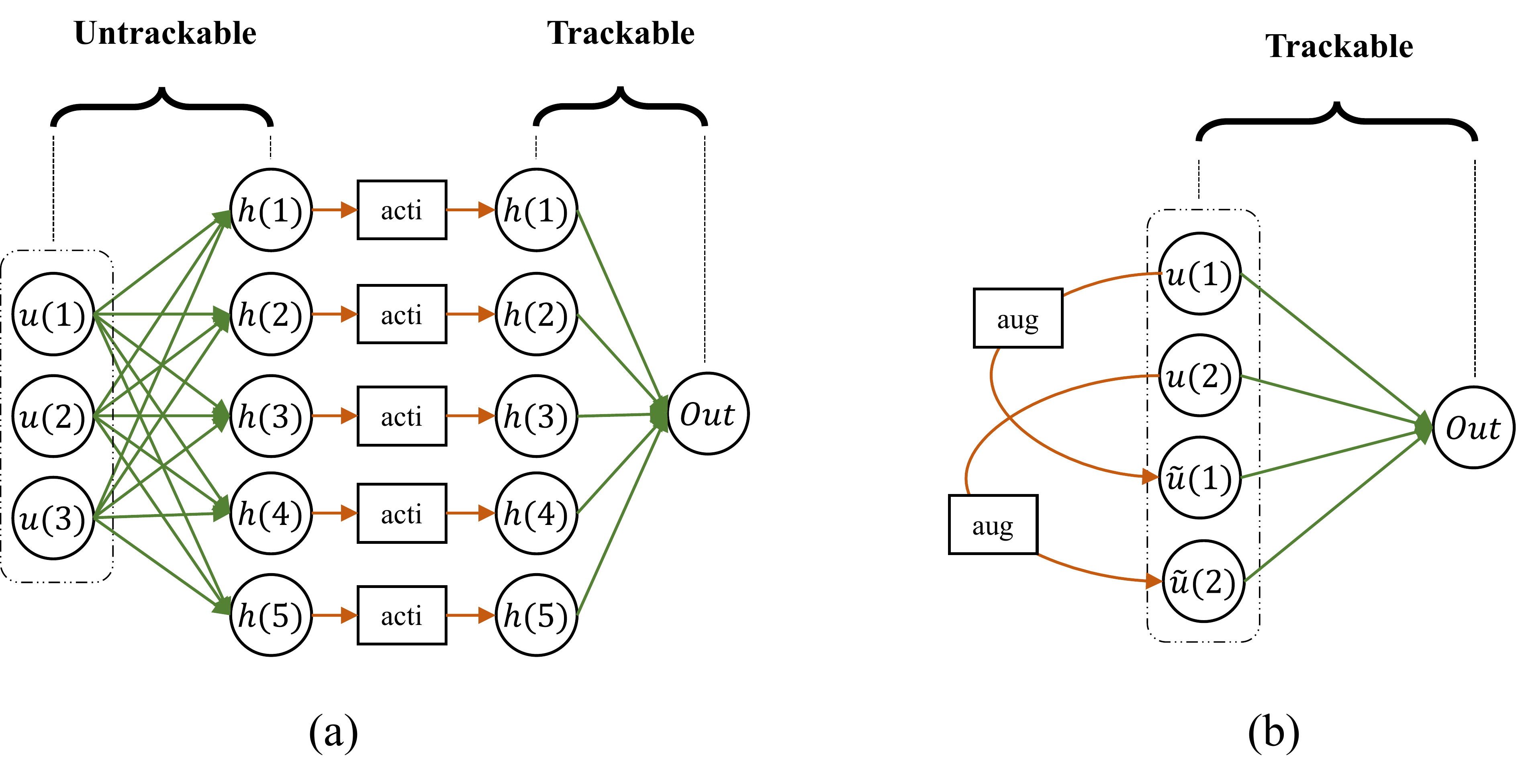}
	\caption{Trackable parameters and Untrackable parameters. 
		Solid green lines represent model parameters, and orange curves represent non-parametric operations. 
		(a) Conventional deep NN framework. 
		(b) Typical broad learning system framework through data enhancement. }
	\label{track and untrack}
\end{figure}

\section{AOPU:Methodology}

\subsection{Trackable vs. Untrackable}

Parameters within NN that can be decoupled from input data $x$ and computed through an inner product are defined as \textbf{trackable} parameters. Conversely, parameters that cannot be decoupled are classified as \textbf{untrackable} parameters.
\begin{equation}
	\begin{aligned}
		f(x)=&W^Tx=\Braket{W,x}, \quad 		g(x)=&M(x)^Tx=\Braket{M(x),x} \\
	\end{aligned}
\end{equation}
where $W$ represents a parameter matrix and $M(\cdot)$ denotes an operator. According to the definition, $W$ is identified as a trackable parameter, whereas $M(x)$ is considered untrackable.

A significant proportion of parameters in NN are untrackable as depicted in Fig. \ref{track and untrack}. 
This predominance is attributable to the networks' reliance on stacking activation functions to bolster their nonlinear modeling capabilities. 
Proposition \ref{pro:1} indicates that any parameter influenced by an activation function transitions to being input-dependent, thus rendering it untrackable.

\begin{proposition}
	\label{pro:1}
	There \textbf{does not} exist an transition operator $T$ independent of $x$ such that for a given parameter matrix $W$, and $\forall x_1,x_2$, the following equations hold,
	\begin{equation}
		\begin{aligned}
			\textup{acti}(Wx_1)&=T(W)x_1,\quad	\textup{acti}(Wx_2)&=T(W)x_2
		\end{aligned}
		\label{eq-propos1}
	\end{equation}
	Proof is in Appendix \ref{sec:proof1}.
\end{proposition}

\subsection{Natural Gradient vs. Gradient}

\begin{figure}
	\centering
	\includegraphics[width=1.\columnwidth]{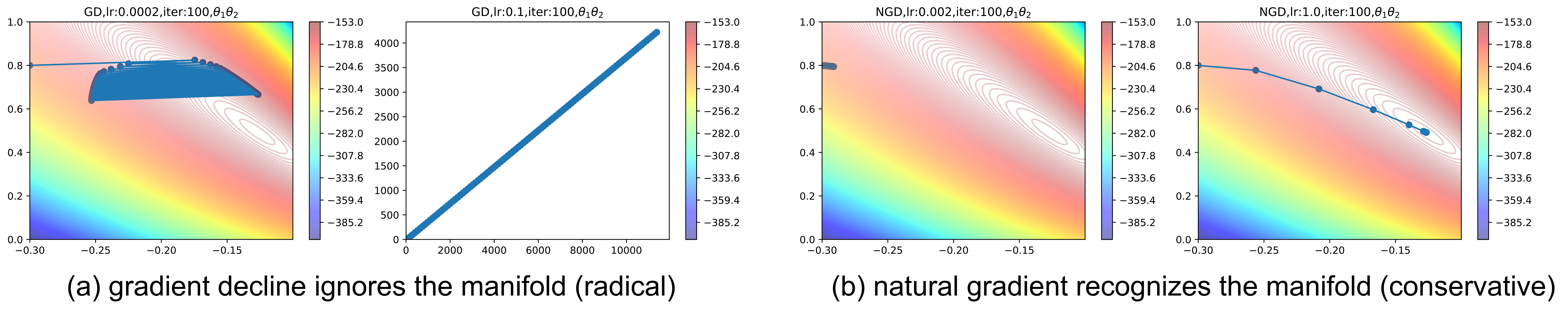}
	\caption{
		Comparison between NGD and GD. Direction matters more than step size (learning rate) in stable convergence.
	}
	\label{ngd vs gd}
\end{figure}

Fig. \ref{ngd vs gd} vividly presents the major difference between NGD and GD using a simple GPR in experiment. 
This GPR had only two parameters, the bias of the mean and the coefficient of the kernel matrix, both constant values. 
We sampled 100 instances from this GPR and updated these two parameters 100 times using these samples. 
It was observed that NG require a higher learning rate, while conventional gradients only need a smaller one. 
The major difference between NG and conventional gradients lies in their directions. 
Conventional gradients ignore the parameter manifold and treat every parameter equally. 
NG, by dividing the gradient by its second derivative, treat sensitive parameters cautiously (low gradient) and non-sensitive parameters boldly (high gradient). 
This adjustment results in different gradient directions and contributes to better convergence.

Nevertheless, the calculation of NG involves considering the inverse of the Fisher Information matrix, thereby introducing computational complexity cubic to the number of parameters, making it entirely infeasible for neural networks. Existing research on NG is almost entirely focused on conventional machine learning, e.g., considering more complex distributions (such as the product of multiple exponential family) for computing NG. Research in the neural network domain on NG mostly centers on second-order optimizers (such as AdamW), which are merely first-order approximations of second-order NG.

\subsection{Network's structure}
\label{sec:network structure}
\textbf{
	We wish to emphasize that within AOPU framework, the truncated gradient of the dual parameters serves as an approximation of the NG of the trackable parameters, while AOPU's structured output approximates the MVE.}
The fidelity of these approximations is measured by the RR, the closer RR is to 1, the more precise the approximation; conversely the closer RR is to 0, the more precision loss occurs. 
Furthermore, it can be demonstrated that the output of AOPU fundamentally differs from traditional neural networks: instead of explicitly modeling a mapping from $x$ to $y$, it implicitly models a mapping from $x$ to $\tilde{x}y$. To ensure that $y$ can be recovered from $\tilde{x}y$, it is imperative that RR equals 1. 
AOPU also guarantees the convergence of the dual parameters if the input-output relationship can be characterized by specific system. 
The proof of above is intricate and comprehensive, and one may refer to Appendix \ref{Network's mechanism}, \ref{convergence analysis} and \ref{mathematic proof} for more detailed information. 
This section focuses on the implementation of AOPU.

AOPU utilize data augmentation to replace stacked activation structures to enhance the nonlinear modeling capabilitie. In such designs, the choice of the data augmentation module forms a crucial model prior. For ease of implementation, AOPU adopts a random weight matrix approach for its data augmentation module \cite{Broad-Learning-System,RVFLNN}. Specifically, suppose the original feature dimension of the input data is $d$, and the defined model hidden dimension size is $h$. Let $\hat{G}$ be a fixed Gaussian initialized random weight matrix, $\hat{G} \in \mathbb{R}^{d,h}$, and for input data $x \in \mathbb{R}^{d,b}$ where $b$ is the mini-batch size, the data augmentation process is as follows,
\begin{equation}
	\begin{aligned}
		\tilde{x}&=\text{concat}[\text{acti}(\hat{G}^{T}x),x]
	\end{aligned}
\end{equation}
Subsequently, for the augmented $\tilde{x}$, the output is processed using the trackable parameter $\tilde{W} \in \mathbb{R}^{(d+h),o}$, where $o$ represents the output dimension and, for simplicity, we assume $o=1$ indicating research into univariate output. The output function is then,
\begin{equation}
	\begin{aligned}
		g(\hat{y}|x)=\tilde{x}^T\tilde{W}
	\end{aligned}
\end{equation}
The optimization of parameter $\tilde{W}$ in AOPU differs from other networks. We introduce a dual parameter $D(\tilde{W})$ to describe this process,
\begin{equation}
	\begin{aligned}
		D(\tilde{W})=\tilde{x}\tilde{x}^T\tilde{W}
	\end{aligned}
\end{equation}
and the loss is computed using the following objective function,
\begin{equation}
	\begin{aligned}
		\mathcal{L} = \frac{1}{b}\sum_{i=1}^{b}\left [y_i-[(\tilde{x}^T\tilde{x})^{-1}\tilde{x}^TD(\tilde{W})]_i\right ]^2
	\end{aligned}
	\label{loss func}
\end{equation}
\begin{figure}
	\centering
	\includegraphics[width=0.6\columnwidth]{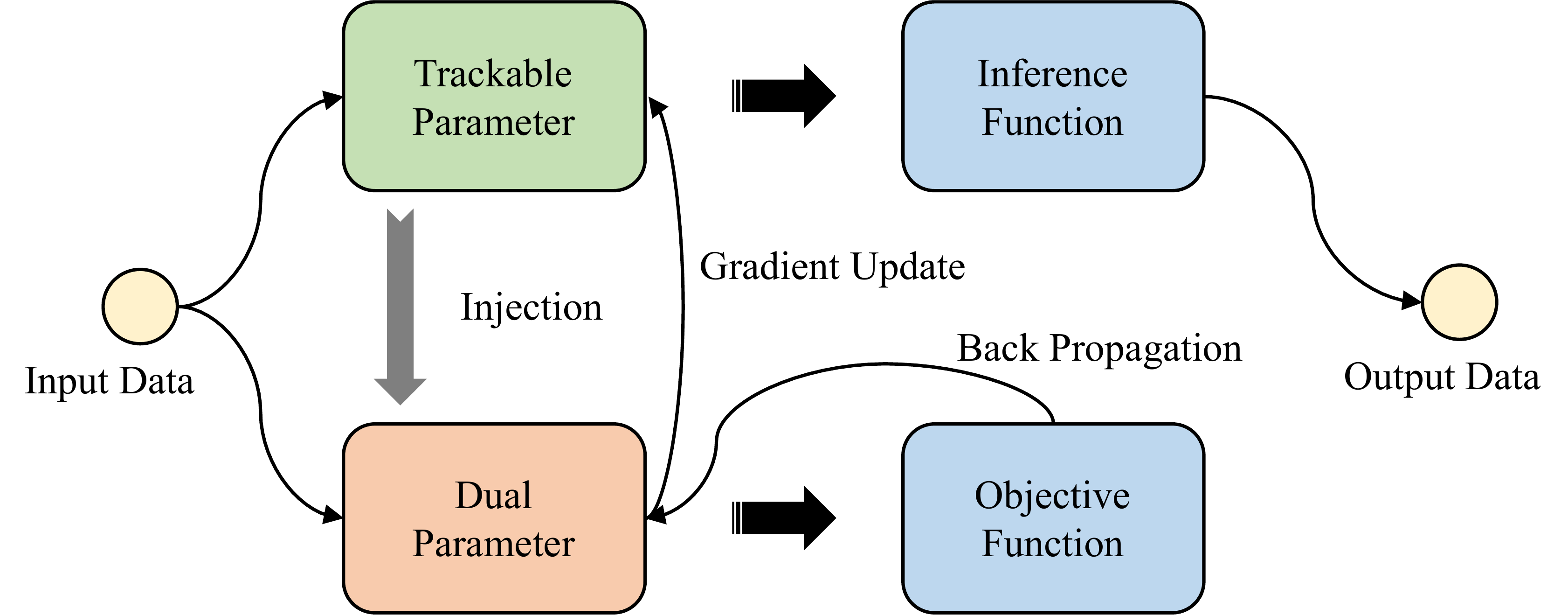}
	\caption{AOPU's data flow schematic. The gradient is backpropagated but truncated at the dual parameter, and this gradient is then used to update the trackable parameter.}
	\label{fig-schematic}
\end{figure}
AOPU innovatively introduces trackable and dual parameters.
The dual parameters are injective representations of trackable parameters, mainly aiding in truncating the gradient backpropagation process. 
During the training process, gradient backpropagation is truncated at $D(\tilde{W})$, and the gradient of the dual parameter updates the original trackable parameter as demonstrated in Fig. \ref{fig-schematic}, thus completing the training of the model.

It is important to note that the training process involves the inversion of a matrix $\tilde{x}^T\tilde{x}$, which is not always invertible, thus introducing numerical computation issue. 
We employ the reciprocals of the positive singular values to circumvent the solvability issues that arise when an inverse does not exist. However, this approach introduces significant computational precision loss, which in turn prompts a thorough analysis of RR.

We define the metric RR to represent the ratio of the rank of $\tilde{x}$ to its batch size. Clearly, RR is a value between $[0,1]$, and as RR approaches 1, the process of approximating the inverse of $\tilde{x}^T\tilde{x}$ through eigenvalue decomposition becomes more accurate (owing to the presence of more reciprocals of eigenvalues). When RR equals 1, $\tilde{x}^T\tilde{x}$ is an invertible matrix; conversely, the smaller the RR, the less stable the model's numerical computations and likely poorer performance.

\section{Experiments and Analysis}
\label{sec-experiment and anlysis}

In this section, we detail the experimental results of the AOPU model, analyze the impact of hyperparameters on AOPU, its robustness regarding changes in hyperparameters, its advantages over other comparative algorithms, and some inherent limitations of the model. 
Comprehensive and detailed experiments and comparisons have been conducted on two publicly available chemical process datasets, Debutanizer and Sulfur Recovery Unit (SRU).
For more information of the dataset please refer to Appendix \ref{data descrip}.
\subsection{Baselines}
We choose seven different NN models as baselines: Autoformer, Informer, DNN, SDAE, SVAE, LSTM, and RVFLNN, covering four major domains including RNN-based networks, auto-encoder-based networks, attention-based networks, and MLP-based networks. Notably, all baseline models except RVFLNN and AOPU operate solely within the latent space, meaning that there are linear transformations mapping the input data to the latent space and from the latent space to the output space before and after the baseline models. This approach is designed to better control the model size.

\subsection{Experiment Implementation}

Apart from AOPU, which is trained using the approximated minimum variance estimation loss function as previously described, all other deep learning algorithms are trained using the Mean Squared Error (MSE) loss. AOPU's learning rate for gradient updates is set at 1.0, while for all other deep learning algorithms, it is set at 0.005, with the Adam optimizer used for gradient updates. The learning rates of all models remain static throughout the training process. 
The experimental setup differs based on the requirements of various models regarding input dimensions. Models such as Autoformer, Informer, and LSTM necessitate an input that includes an additional dimension for 'sequence length'. This dimension is preserved as part of the input structure for these models. Conversely, models like DNN, SDAE, SVAE, AOPU, and RVFLNN do not require this additional dimension. For these models, the sequence length and input dimensions are combined and flattened to serve as the feature dimensions in the input. 
AOPU's latent space size is set at 2048. Autoformer, Informer, SDAE, and SVAE utilize two layers each for their encoder and decoder layers; LSTM uses two layers of LSTM layers; RVFLNN and AOPU share identical settings. All models except AOPU and RVFLNN have their latent space sizes set at 16 to ensure the trainable parameters size across all models are comparable.

\subsection{Main Result}

\subsubsection{How certain we are about the inverse}
According to the previous discussion, the existence of the inverse of $\tilde{x}^T\tilde{x}$ is crucial as it does not only impact the numerical stability of the model but also directly determines whether it is possible to recover $y$ from the approximated mapping relationship $\tilde{x} \rightarrow \tilde{x}y$. Clearly, the input feature dimensions $d+h$ and the batch size $b$ significantly affect whether the inverse of $\tilde{x}^T\tilde{x}$ exists. Specifically, the larger the batch size, the more columns $\tilde{x}$ has, and the less likely $\tilde{x}$ is to be column-full-rank; conversely, the longer the sequence length and the larger the input feature dimensions, the more likely $\tilde{x}$ is to be linearly independent and thus column-full-rank.  From the following experimental results, it will be clearly observed the impact of batch size and sequence length on RR.

\begin{figure*}
	\centering
	\begin{adjustbox}{center}
		\includegraphics[width=.7\columnwidth]{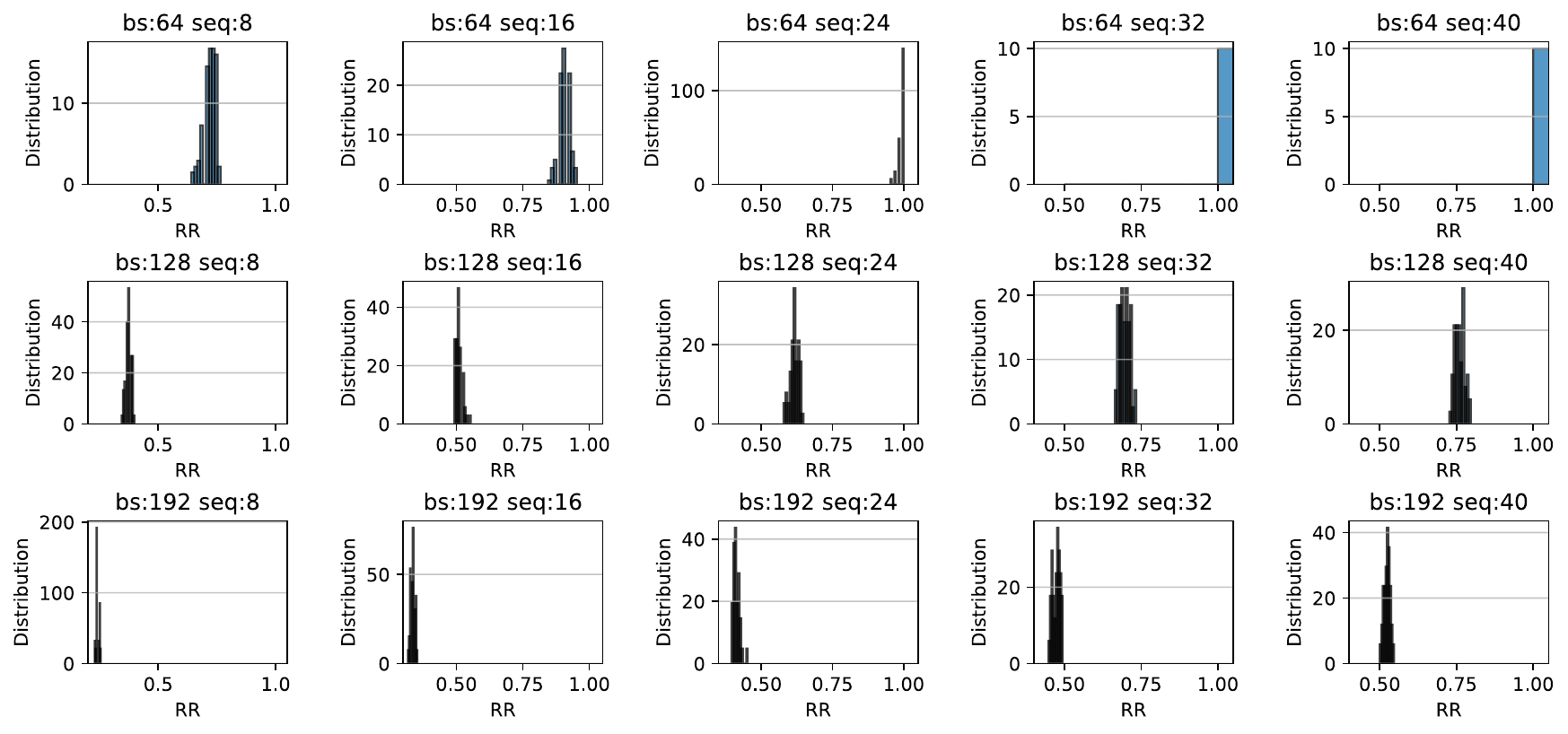}
	\end{adjustbox}
	\caption{Histogram of the frequency distribution of RR on SRU dataset under varying batch sizes and sequence length settings. }
	\label{sru distribution}
\end{figure*}

Fig. \ref{sru distribution} shows the distribution of the RR for the AOPU model across various batch sizes and sequence length combinations on the SRU dataset, where \textbf{bs} stands for batch size and \textbf{seq} for sequence length. The experimental results align with the previous analysis: increasing the batch size with a fixed sequence length significantly decreases the RR distribution, whereas increasing the sequence length with a fixed batch size significantly increases it.

\begin{figure*}
	\centering
	\begin{adjustbox}{center}
		\includegraphics[width=.7\columnwidth]{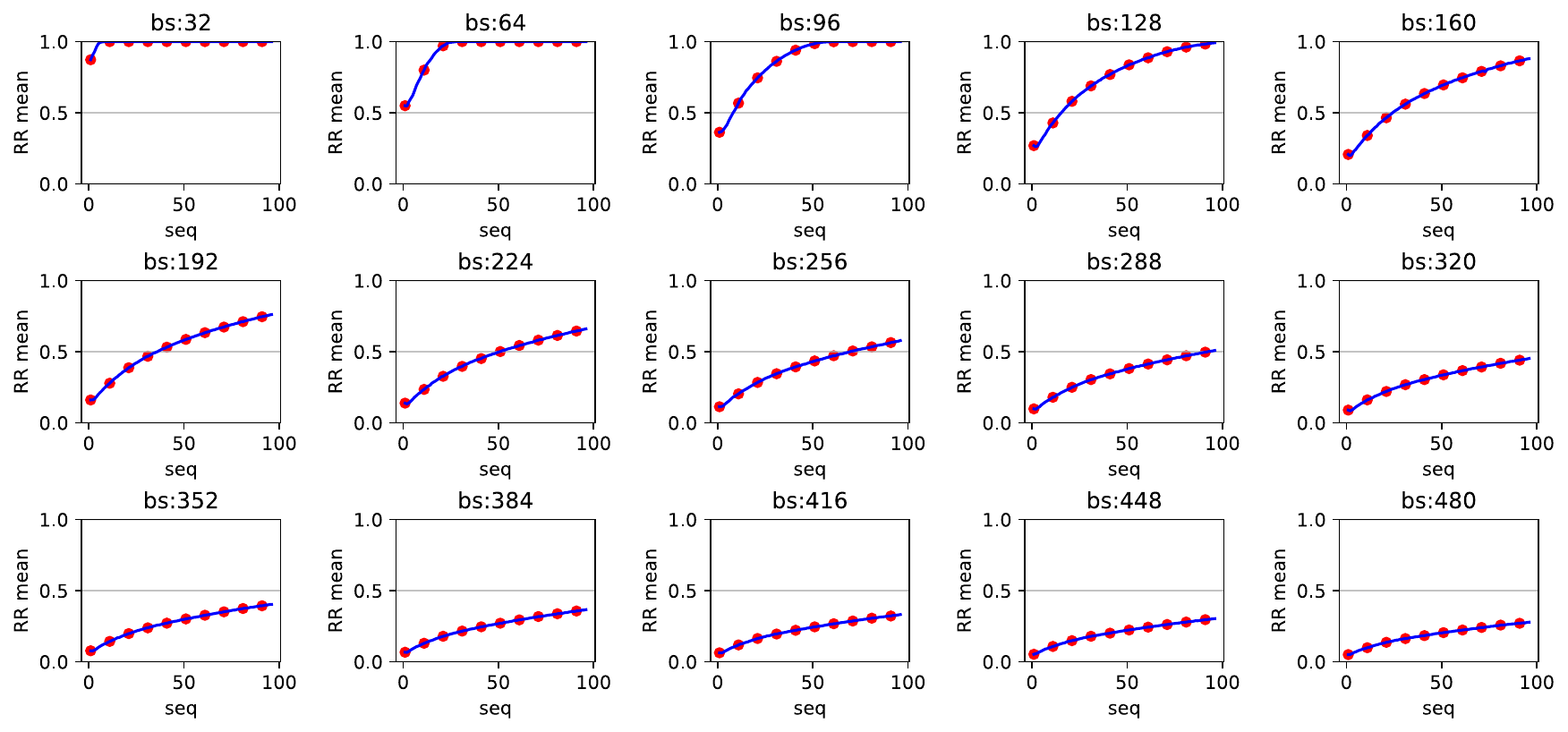}
	\end{adjustbox}
	\caption{Curve of the mean of RR distribution on SRU dataset under varying batch sizes and sequence length settings. }
	\label{sru trend}
\end{figure*}

Fig. \ref{sru trend} shows the mean values of RR distribution changing with sequence length under different batch size settings, marked by red circles at every ten data points. Clearly, the experimental results shown in the figure corroborate our analysis that with increasing batch size, the curve's slope becomes flatter, indicating the model's decreasing sensitivity to changes in sequence length. Compared to Fig. \ref{sru distribution}, Fig. \ref{sru trend} provides additional insights into the mean values of the RR distribution relative to sequence length and batch size, offering a more comprehensive insight for subsequent experimental interpretations.
Results of the RR study on the Debutanizer are listed in Appendix \ref{supp figure}.

\begin{figure*}
	\centering
	\begin{adjustbox}{center}
		\includegraphics[width=.7\columnwidth]{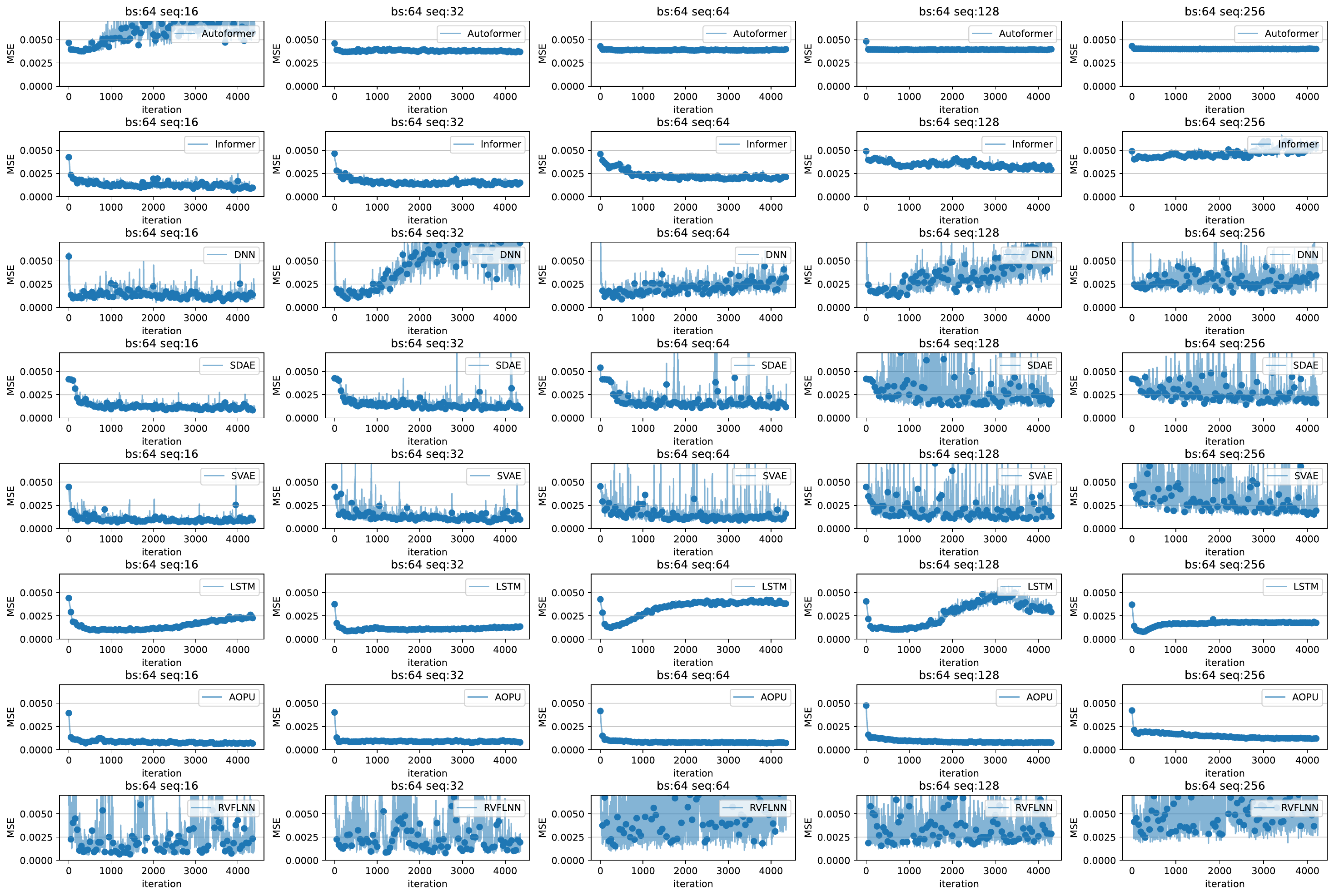}
	\end{adjustbox}
	\caption{Curves of SRU validation loss changes with training iteration for different models with a fixed batch size of 64 at different sequence length settings.  The curves are shown in translucent blue, with a solid blue circle labeled on the curve every 50 iterations.}
	\label{sru bs64}
\end{figure*}

\subsubsection{Is the training stable}  
Stability is a crucial characteristic for the online deployment of deep learning models in actual production processes. Specifically, the incremental updates to model parameters following the observation of new mini-batch data should have a smooth impact on model performance. However, experimental results indicate that most networks in the soft sensor field fail to achieve stable convergence. Fig. \ref{sru bs64} provides a detailed display of how the MSE metrics for different networks change with training iterations on the SRU validation dataset, with blue solid circles marked every 50 iterations. It is evident that all models, except Autoformer, Informer, LSTM, and AOPU, exhibit significant performance fluctuations as training iterations progress. The density of the blue solid circles can to some extent represent the likelihood of corresponding performance fluctuations.

It can be observed that the SDAE and SVAE networks, despite experiencing significant fluctuations in validation performance (indicated by large fluctuations in the curve), are mostly stable (as shown by the blue circles concentrated below the curve). In contrast, the DNN and RVFLNN networks have relatively unstable convergence (indicated by blue circles evenly distributed above and below the curve). Although Autoformer and Informer have relatively stable convergence dynamics, their performance is relatively poor. Specifically, Autoformer consistently converges to a bad output, whereas Informer can effectively learn under identical settings but is sensitive to changes in seq, which can lead to model performance collapse. The convergence process of LSTM is relatively stable, partly explaining why it is a widely adopted baseline in the field of time series analysis; however, LSTM is significantly prone to overfitting and its performance is not outstanding.

In contrast to all other network, AOPU exhibits exceptionally impressive performance. AOPU demonstrates very stable and rapid convergence, with almost no fluctuations in performance as training iterations progress. Furthermore, AOPU is less sensitive to changes in hyperparameters and does not exhibit significant overfitting, making it a truly reliable and deployable NN model in production processes. We also notice that with bigger batch size setting, the fluctuations will be mitigated. For comparative training dynamics under other batch size settings, refer to Appendix \ref{supp figure}.

\subsubsection{Quantitative analysis}
\label{Comparsion result}

\begin{table*}
	\caption{The evaluation metrics for different models under different seq settings on Debutanizer and SRU dataset with bs set to 64 and best training epoch recorded}
	\label{tab:comprehensive best experiments}
	\centering
	{\scriptsize
		\setlength{\tabcolsep}{5pt}
		\begin{adjustbox}{center}
			\begin{threeparttable}
				\begin{tabular}{c|c|ccc|ccc}
					\toprule
					\multicolumn{2}{c}{Model} & \multicolumn{6}{c}{Dataset \& Metric} \\
					\midrule
					\multirow{2}{*}{Seq} & \multirow{2}{*}{Name} & \multicolumn{3}{c|}{Debutanizer} & \multicolumn{3}{c}{SRU} \\
					& & MSE & MAPE & R$^2$ & MSE & MAPE & R$^2$ \\
					\midrule
					\multirow{8}{*}{16} 
					& Autoformer & 0.0314$_{\pm0.0017}$ & 177.2$_{\pm24.56}$ & 0.1588$_{\pm0.0481}$ & 0.00312$_{\pm0.00037}$ & 0.3295$_{\pm0.0310}$ & 0.2079$_{\pm0.0957}$ \\
					& Informer & 0.0165$_{\pm0.0023}$ & 130.9$_{\pm19.26}$ & 0.5574$_{\pm0.0641}$ & 0.00090$_{\pm0.00011}$ & 0.1579$_{\pm0.0117}$ & 0.7715$_{\pm0.0294}$ \\
					& DNN & 0.0146$_{\pm0.0016}$ & 121.9$_{\pm30.82}$ & 0.6076$_{\pm0.0440}$ & 0.00081$_{\pm0.00017}$ & 0.1645$_{\pm0.0249}$ & 0.7944$_{\pm0.0441}$ \\
					& SDAE & 0.0134$_{\pm0.0021}$ & 136.1$_{\pm13.79}$ & 0.6406$_{\pm0.0573}$ & 0.00080$_{\pm0.00006}$ & 0.1453$_{\pm0.0102}$ & 0.7967$_{\pm0.0154}$ \\
					& SVAE & 0.0138$_{\pm0.0013}$ & 128.8$_{\pm30.74}$ & 0.6294$_{\pm0.0352}$ & 0.00078$_{\pm0.00004}$ & 0.1214$_{\pm0.0091}$ & 0.7992$_{\pm0.0102}$ \\
					& LSTM & 0.0259$_{\pm0.0027}$ & 161.2$_{\pm50.17}$ & 0.3058$_{\pm0.0741}$ & 0.00088$_{\pm0.00011}$ & 0.1445$_{\pm0.0106}$ & 0.7750$_{\pm0.0292}$ \\
					& AOPU & 0.0149$_{\pm0.0009}$ & 200.5$_{\pm9.277}$ & 0.5992$_{\pm0.0243}$ & 0.00072$_{\pm0.00007}$ & 0.1694$_{\pm0.0114}$ & 0.8154$_{\pm0.0190}$ \\
					& RVFLNN & 0.0114$_{\pm0.0019}$ & 78.88$_{\pm36.11}$ & 0.6936$_{\pm0.0534}$ & 0.00074$_{\pm0.00009}$ & 0.1584$_{\pm0.0148}$ & 0.8128$_{\pm0.0249}$ \\
					\noalign{\vskip 3pt}
					\noalign{\vskip -2pt} \cline{1-8}
					\noalign{\vskip 3pt}
					\multirow{8}{*}{24} 
					& Autoformer & 0.0295$_{\pm0.0020}$ & 153.1$_{\pm39.66}$ & 0.2091$_{\pm0.0550}$ & 0.00236$_{\pm0.00030}$ & 0.2685$_{\pm0.0262}$ & 0.4014$_{\pm0.0774}$ \\
					& Informer & 0.0154$_{\pm0.0042}$ & 139.3$_{\pm19.59}$ & 0.5864$_{\pm0.1145}$ & 0.00110$_{\pm0.00022}$ & 0.1670$_{\pm0.0232}$ & 0.7210$_{\pm0.0572}$ \\
					& DNN & 0.0122$_{\pm0.0027}$ & 124.7$_{\pm43.75}$ & 0.6717$_{\pm0.0733}$ & 0.00091$_{\pm0.00014}$ & 0.1671$_{\pm0.0192}$ & 0.7684$_{\pm0.0361}$ \\
					& SDAE & 0.0120$_{\pm0.0023}$ & 144.9$_{\pm25.18}$ & 0.6786$_{\pm0.0631}$ & 0.00076$_{\pm0.00009}$ & 0.1489$_{\pm0.0131}$ & 0.8070$_{\pm0.0234}$ \\
					& SVAE & 0.0127$_{\pm0.0023}$ & 143.0$_{\pm33.10}$ & 0.6599$_{\pm0.0642}$ & 0.00072$_{\pm0.00004}$ & 0.1289$_{\pm0.0143}$ & 0.8149$_{\pm0.0122}$ \\
					& LSTM & 0.0262$_{\pm0.0029}$ & 192.2$_{\pm48.06}$ & 0.2961$_{\pm0.0795}$ & 0.00091$_{\pm0.00014}$ & 0.1420$_{\pm0.0122}$ & 0.7672$_{\pm0.0364}$ \\
					& AOPU & 0.0133$_{\pm0.0006}$ & 178.3$_{\pm11.16}$ & 0.6421$_{\pm0.0181}$ & 0.00080$_{\pm0.00004}$ & 0.1868$_{\pm0.0063}$ & 0.7965$_{\pm0.0110}$ \\
					& RVFLNN & 0.0109$_{\pm0.0012}$ & 67.40$_{\pm26.06}$ & 0.7070$_{\pm0.0338}$ & 0.00079$_{\pm0.00011}$ & 0.1638$_{\pm0.0168}$ & 0.7986$_{\pm0.0297}$ \\
					\noalign{\vskip 3pt}
					\noalign{\vskip -2pt} \cline{1-8}
					\noalign{\vskip 3pt}
					\multirow{8}{*}{32} 
					& Autoformer & 0.0307$_{\pm0.0018}$ & 199.3$_{\pm31.82}$ & 0.1770$_{\pm0.0495}$ & 0.00222$_{\pm0.00022}$ & 0.2730$_{\pm0.0291}$ & 0.4366$_{\pm0.0573}$ \\
					& Informer & 0.0237$_{\pm0.0049}$ & 119.4$_{\pm18.42}$ & 0.3648$_{\pm0.1325}$ & 0.00131$_{\pm0.00020}$ & 0.1849$_{\pm0.0164}$ & 0.6679$_{\pm0.0522}$ \\
					& DNN & 0.0120$_{\pm0.0027}$ & 93.96$_{\pm45.53}$ & 0.6779$_{\pm0.0735}$ & 0.00095$_{\pm0.00014}$ & 0.1693$_{\pm0.0149}$ & 0.7589$_{\pm0.0374}$ \\
					& SDAE & 0.0111$_{\pm0.0020}$ & 131.9$_{\pm27.59}$ & 0.686$_{\pm0.0539}$ & 0.00080$_{\pm0.00005}$ & 0.1534$_{\pm0.0124}$ & 0.7959$_{\pm0.0145}$ \\
					& SVAE & 0.0133$_{\pm0.0024}$ & 143.9$_{\pm33.72}$ & 0.6419$_{\pm0.0646}$ & 0.00078$_{\pm0.00007}$ & 0.1300$_{\pm0.0118}$ & 0.8010$_{\pm0.0182}$ \\
					& LSTM & 0.0257$_{\pm0.0029}$ & 180.1$_{\pm48.29}$ & 0.3097$_{\pm0.0785}$ & 0.00094$_{\pm0.00009}$ & 0.1488$_{\pm0.0084}$ & 0.7603$_{\pm0.0234}$ \\
					& AOPU & 0.0119$_{\pm0.0005}$ & 155.6$_{\pm4.827}$ & 0.6791$_{\pm0.0155}$ & 0.00076$_{\pm0.00003}$ & 0.1847$_{\pm0.0048}$ & 0.8063$_{\pm0.0078}$ \\
					& RVFLNN & 0.0121$_{\pm0.0016}$ & 57.82$_{\pm34.15}$ & 0.6751$_{\pm0.0445}$ & 0.00091$_{\pm0.00012}$ & 0.1725$_{\pm0.0129}$ & 0.7679$_{\pm0.0308}$ \\
					\noalign{\vskip 3pt}
					\noalign{\vskip -2pt} \cline{1-8}
					\noalign{\vskip 3pt}
					\multirow{8}{*}{40} 
					& Autoformer & 0.0307$_{\pm0.0023}$ & 199.4$_{\pm21.66}$ & 0.1771$_{\pm0.0642}$ & 0.00245$_{\pm0.00020}$ & 0.2879$_{\pm0.0210}$ & 0.3790$_{\pm0.0509}$ \\
					& Informer & 0.0237$_{\pm0.0059}$ & 119.4$_{\pm32.86}$ & 0.3648$_{\pm0.1597}$ & 0.00168$_{\pm0.00054}$ & 0.2205$_{\pm0.0334}$ & 0.5723$_{\pm0.1369}$ \\
					& DNN & 0.0120$_{\pm0.0029}$ & 93.97$_{\pm42.94}$ & 0.678$_{\pm0.0792}$ & 0.00099$_{\pm0.00017}$ & 0.1784$_{\pm0.0191}$ & 0.7489$_{\pm0.0445}$ \\
					& SDAE & 0.0111$_{\pm0.0023}$ & 134.4$_{\pm36.79}$ & 0.7009$_{\pm0.0618}$ & 0.00081$_{\pm0.00007}$ & 0.1546$_{\pm0.0119}$ & 0.7938$_{\pm0.0180}$ \\
					& SVAE & 0.0116$_{\pm0.0025}$ & 127.2$_{\pm31.86}$ & 0.6868$_{\pm0.0687}$ & 0.00082$_{\pm0.00012}$ & 0.1383$_{\pm0.0162}$ & 0.7898$_{\pm0.0310}$ \\
					& LSTM & 0.0261$_{\pm0.0039}$ & 197.2$_{\pm56.01}$ & 0.2994$_{\pm0.1059}$ & 0.00093$_{\pm0.00011}$ & 0.1479$_{\pm0.0093}$ & 0.7638$_{\pm0.0284}$ \\
					& AOPU & 0.0110$_{\pm0.0004}$ & 143.4$_{\pm6.977}$ & 0.7030$_{\pm0.0121}$ & 0.00071$_{\pm0.00001}$ & 0.1785$_{\pm0.0034}$ & 0.8193$_{\pm0.0048}$ \\
					& RVFLNN & 0.0148$_{\pm0.0013}$ & 53.59$_{\pm30.26}$ & 0.6013$_{\pm0.0373}$ & 0.00093$_{\pm0.00013}$ & 0.1761$_{\pm0.0170}$ & 0.7644$_{\pm0.0337}$ \\
					\noalign{\vskip 3pt}
					\noalign{\vskip -2pt} \cline{1-8}
					\noalign{\vskip 3pt}
					\multirow{8}{*}{48}
					& Autoformer & 0.0321$_{\pm0.0022}$ & 189.4$_{\pm24.87}$ & 0.1390$_{\pm0.0614}$ & 0.00249$_{\pm0.00015}$ & 0.2879$_{\pm0.0220}$ & 0.3676$_{\pm0.0380}$ \\
					& Informer & 0.0247$_{\pm0.0069}$ & 122.6$_{\pm36.11}$ & 0.3385$_{\pm0.1865}$ & 0.00190$_{\pm0.00043}$ & 0.2270$_{\pm0.0386}$ & 0.5186$_{\pm0.1092}$ \\
					& DNN & 0.0114$_{\pm0.0027}$ & 110.0$_{\pm42.75}$ & 0.6941$_{\pm0.0732}$ & 0.00105$_{\pm0.00013}$ & 0.1796$_{\pm0.0168}$ & 0.7332$_{\pm0.0344}$ \\
					& SDAE & 0.0105$_{\pm0.0014}$ & 128.5$_{\pm16.98}$ & 0.7182$_{\pm0.0400}$ & 0.00083$_{\pm0.00005}$ & 0.1564$_{\pm0.0106}$ & 0.7890$_{\pm0.0133}$ \\
					& SVAE & 0.0105$_{\pm0.0017}$ & 116.5$_{\pm34.32}$ & 0.7180$_{\pm0.0473}$ & 0.00087$_{\pm0.00010}$ & 0.1389$_{\pm0.0098}$ & 0.7789$_{\pm0.0256}$ \\
					& LSTM & 0.0268$_{\pm0.0036}$ & 185.6$_{\pm63.68}$ & 0.2813$_{\pm0.0982}$ & 0.00091$_{\pm0.00012}$ & 0.1471$_{\pm0.0099}$ & 0.7691$_{\pm0.0312}$ \\
					& AOPU & 0.0107$_{\pm0.0003}$ & 144.9$_{\pm7.492}$ & 0.7118$_{\pm0.0087}$ & 0.00070$_{\pm0.00001}$ & 0.1796$_{\pm0.0019}$ & 0.8209$_{\pm0.0037}$ \\
					& RVFLNN & 0.0157$_{\pm0.0026}$ & 60.11$_{\pm29.00}$ & 0.5792$_{\pm0.0717}$ & 0.00100$_{\pm0.00016}$ & 0.1833$_{\pm0.0205}$ & 0.7454$_{\pm0.0412}$ \\
					\bottomrule
				\end{tabular}
			\end{threeparttable}
		\end{adjustbox}
	}
\end{table*}

\begin{table*}
	\caption{The evaluation metrics for different models under different seq settings on Debutanizer and SRU dataset with bs set to 64 and no best training epoch recorded}
	\label{tab:comprehensive hard experiments}
	\centering
	{\scriptsize
		\setlength{\tabcolsep}{5pt}
		\begin{adjustbox}{center}
			\begin{threeparttable}
				\begin{tabular}{c|c|ccc|ccc}
					\toprule
					\multicolumn{2}{c}{Model} & \multicolumn{6}{c}{Dataset \& Metric} \\
					\midrule
					\multirow{2}{*}{Seq} & \multirow{2}{*}{Name} & \multicolumn{3}{c|}{Debutanizer} & \multicolumn{3}{c}{SRU} \\
					& & MSE & MAPE & R$^2$ & MSE & MAPE & R$^2$ \\
					\midrule
					\multirow{8}{*}{16}
					& Autoformer & 0.0942$_{\pm0.0370}$ & 217.0$_{\pm78.49}$ & -1.5240$_{\pm0.9912}$ & 0.00836$_{\pm0.00426}$ & 0.5362$_{\pm0.1755}$ & -1.118$_{\pm1.0810}$ \\
					& Informer & 0.0283$_{\pm0.0070}$ & 140.8$_{\pm30.66}$ & 0.2414$_{\pm0.1885}$ & 0.00122$_{\pm0.00036}$ & 0.1750$_{\pm0.0226}$ & 0.6897$_{\pm0.0918}$ \\
					& DNN & 0.0215$_{\pm0.0028}$ & 161.8$_{\pm46.14}$ & 0.4233$_{\pm0.0760}$ & 0.00172$_{\pm0.00073}$ & 0.2238$_{\pm0.0410}$ & 0.5639$_{\pm0.1850}$ \\
					& SDAE & 0.0217$_{\pm0.0036}$ & 144.6$_{\pm43.52}$ & 0.4184$_{\pm0.0977}$ & 0.00113$_{\pm0.00017}$ & 0.1559$_{\pm0.0211}$ & 0.7114$_{\pm0.0453}$ \\
					& SVAE & 0.0217$_{\pm0.0061}$ & 161.6$_{\pm53.46}$ & 0.4166$_{\pm0.1644}$ & 0.00113$_{\pm0.00044}$ & 0.1469$_{\pm0.0329}$ & 0.7121$_{\pm0.1131}$ \\
					& LSTM & 0.0521$_{\pm0.0152}$ & 215.5$_{\pm102.7}$ & -0.3956$_{\pm0.4094}$ & 0.00251$_{\pm0.00112}$ & 0.2055$_{\pm0.0325}$ & 0.3634$_{\pm0.2844}$ \\
					& AOPU & 0.0215$_{\pm0.0007}$ & 206.6$_{\pm9.059}$ & 0.4239$_{\pm0.0211}$ & 0.00098$_{\pm0.00013}$ & 0.1963$_{\pm0.0132}$ & 0.7518$_{\pm0.0336}$ \\
					& RVFLNN & 0.0329$_{\pm0.0391}$ & 107.1$_{\pm40.42}$ & 0.1171$_{\pm1.0470}$ & 0.00171$_{\pm0.00112}$ & 0.2540$_{\pm0.0867}$ & 0.5652$_{\pm0.2853}$ \\
					\noalign{\vskip 3pt}
					\noalign{\vskip -2pt} \cline{1-8}
					\noalign{\vskip 3pt}
					\multirow{8}{*}{24} 
					& Autoformer & 0.0969$_{\pm0.0293}$ & 315.7$_{\pm113.7}$ & -1.5980$_{\pm0.7864}$ & 0.00457$_{\pm0.00145}$ & 0.3901$_{\pm0.0764}$ & -0.1590$_{\pm0.3682}$ \\
					& Informer & 0.0222$_{\pm0.0053}$ & 134.9$_{\pm29.93}$ & 0.4047$_{\pm0.1432}$ & 0.00146$_{\pm0.00047}$ & 0.1878$_{\pm0.0280}$ & 0.6280$_{\pm0.1201}$ \\
					& DNN & 0.0204$_{\pm0.0036}$ & 159.7$_{\pm27.21}$ & 0.4518$_{\pm0.0972}$ & 0.00281$_{\pm0.00126}$ & 0.2760$_{\pm0.0699}$ & 0.2879$_{\pm0.3206}$ \\
					& SDAE & 0.0206$_{\pm0.0069}$ & 152.2$_{\pm44.43}$ & 0.4470$_{\pm0.1873}$ & 0.00118$_{\pm0.00023}$ & 0.1688$_{\pm0.0219}$ & 0.6998$_{\pm0.0590}$ \\
					& SVAE & 0.0217$_{\pm0.0048}$ & 152.6$_{\pm33.46}$ & 0.4165$_{\pm0.1306}$ & 0.00106$_{\pm0.00034}$ & 0.1582$_{\pm0.0248}$ & 0.7286$_{\pm0.0873}$ \\
					& LSTM & 0.0544$_{\pm0.0146}$ & 226.9$_{\pm129.5}$ & -0.4574$_{\pm0.3932}$ & 0.00313$_{\pm0.00141}$ & 0.1878$_{\pm0.0337}$ & 0.6280$_{\pm0.3582}$ \\
					& AOPU & 0.0185$_{\pm0.0005}$ & 193.5$_{\pm5.519}$ & 0.5040$_{\pm0.0136}$ & 0.00093$_{\pm0.00009}$ & 0.2022$_{\pm0.0099}$ & 0.7628$_{\pm0.0232}$ \\
					& RVFLNN & 0.0338$_{\pm0.0470}$ & 135.0$_{\pm86.13}$ & 0.0945$_{\pm1.2610}$ & 0.00507$_{\pm0.00725}$ & 0.3854$_{\pm0.2346}$ & -0.2849$_{\pm1.8360}$ \\
					\noalign{\vskip 3pt}
					\noalign{\vskip -2pt} \cline{1-8}
					\noalign{\vskip 3pt}
					\multirow{8}{*}{32}
					& Autoformer & 0.0583$_{\pm0.0194}$ & 217.3$_{\pm48.84}$ & -0.5629$_{\pm0.5216}$ & 0.00334$_{\pm0.00101}$ & 0.3279$_{\pm0.0407}$ & 0.1536$_{\pm0.2573}$ \\
					& Informer & 0.0359$_{\pm0.0134}$ & 135.4$_{\pm43.65}$ & 0.0371$_{\pm0.3612}$ & 0.00193$_{\pm0.00064}$ & 0.2163$_{\pm0.0331}$ & 0.5091$_{\pm0.1637}$ \\
					& DNN & 0.0206$_{\pm0.0049}$ & 173.2$_{\pm44.64}$ & 0.4459$_{\pm0.1316}$ & 0.00278$_{\pm0.00146}$ & 0.2780$_{\pm0.0858}$ & 0.2949$_{\pm0.3695}$ \\
					& SDAE & 0.0211$_{\pm0.0045}$ & 179.3$_{\pm61.21}$ & 0.4342$_{\pm0.1228}$ & 0.00110$_{\pm0.00032}$ & 0.1812$_{\pm0.0253}$ & 0.7209$_{\pm0.0824}$ \\
					& SVAE & 0.0239$_{\pm0.0083}$ & 151.3$_{\pm46.68}$ & 0.3598$_{\pm0.2234}$ & 0.00102$_{\pm0.00049}$ & 0.1669$_{\pm0.0361}$ & 0.7397$_{\pm0.1256}$ \\
					& LSTM & 0.0519$_{\pm0.0204}$ & 243.7$_{\pm75.24}$ & -0.3903$_{\pm0.5471}$ & 0.00284$_{\pm0.00132}$ & 0.2186$_{\pm0.0513}$ & 0.2803$_{\pm0.3348}$ \\
					& AOPU & 0.0171$_{\pm0.0003}$ & 191.0$_{\pm4.483}$ & 0.5396$_{\pm0.0105}$ & 0.00085$_{\pm0.00004}$ & 0.1945$_{\pm0.0070}$ & 0.7833$_{\pm0.0118}$ \\
					& RVFLNN & 0.0305$_{\pm0.0340}$ & 86.84$_{\pm46.24}$ & 0.1809$_{\pm0.9113}$ & 0.00741$_{\pm0.01854}$ & 0.4012$_{\pm0.4214}$ & -0.8785$_{\pm4.6942}$ \\
					\noalign{\vskip 3pt}
					\noalign{\vskip -2pt} \cline{1-8}
					\noalign{\vskip 3pt}
					\multirow{8}{*}{40} 
					& Autoformer & 0.0462$_{\pm0.0136}$ & 213.2$_{\pm62.30}$ & -0.2396$_{\pm0.3648}$ & 0.00341$_{\pm0.00077}$ & 0.3528$_{\pm0.0453}$ & 0.1343$_{\pm0.1970}$ \\
					& Informer & 0.0194$_{\pm0.0059}$ & 119.5$_{\pm23.91}$ & -0.1124$_{\pm0.5220}$ & 0.00201$_{\pm0.00049}$ & 0.2283$_{\pm0.0285}$ & 0.4911$_{\pm0.1259}$ \\
					& DNN & 0.0182$_{\pm0.0033}$ & 166.4$_{\pm36.94}$ & 0.5109$_{\pm0.0892}$ & 0.00422$_{\pm0.00360}$ & 0.3376$_{\pm0.1262}$ & -0.0703$_{\pm0.9134}$ \\
					& SDAE & 0.0223$_{\pm0.0053}$ & 158.9$_{\pm37.77}$ & 0.4028$_{\pm0.1430}$ & 0.00140$_{\pm0.00096}$ & 0.1987$_{\pm0.0462}$ & 0.6455$_{\pm0.2437}$ \\
					& SVAE & 0.0214$_{\pm0.0063}$ & 147.5$_{\pm43.71}$ & 0.4259$_{\pm0.1705}$ & 0.00118$_{\pm0.00070}$ & 0.1699$_{\pm0.0296}$ & 0.7001$_{\pm0.1790}$ \\
					& LSTM & 0.0534$_{\pm0.0211}$ & 219.6$_{\pm80.80}$ & -0.4307$_{\pm0.5669}$ & 0.00391$_{\pm0.00244}$ & 0.2460$_{\pm0.0584}$ & 0.0090$_{\pm0.6181}$ \\
					& AOPU & 0.0156$_{\pm0.0003}$ & 179.0$_{\pm4.713}$ & 0.5804$_{\pm0.0105}$ & 0.00076$_{\pm0.00004}$ & 0.1843$_{\pm0.0052}$ & 0.8069$_{\pm0.0101}$ \\
					& RVFLNN & 0.0267$_{\pm0.0130}$ & 88.96$_{\pm30.42}$ & 0.2844$_{\pm0.3489}$ & 0.00338$_{\pm0.00277}$ & 0.3133$_{\pm0.1144}$ & 0.1442$_{\pm0.7023}$ \\
					\noalign{\vskip 3pt}
					\noalign{\vskip -2pt} \cline{1-8}
					\noalign{\vskip 3pt}
					\multirow{8}{*}{48}
					& Autoformer & 0.0492$_{\pm0.0122}$ & 226.2$_{\pm46.25}$ & -0.3187$_{\pm0.3287}$ & 0.00338$_{\pm0.00044}$ & 0.3655$_{\pm0.0331}$ & 0.1440$_{\pm0.1124}$ \\
					& Informer & 0.0440$_{\pm0.0175}$ & 123.7$_{\pm40.20}$ & -0.1798$_{\pm0.4709}$ & 0.00248$_{\pm0.00072}$ & 0.2389$_{\pm0.0309}$ & 0.3712$_{\pm0.1826}$ \\
					& DNN & 0.0194$_{\pm0.0043}$ & 167.9$_{\pm37.16}$ & 0.4799$_{\pm0.1166}$ & 0.00393$_{\pm0.00219}$ & 0.3284$_{\pm0.1002}$ & 0.0046$_{\pm0.5544}$ \\
					& SDAE & 0.0212$_{\pm0.0066}$ & 153.0$_{\pm44.14}$ & 0.4316$_{\pm0.1772}$ & 0.00120$_{\pm0.00028}$ & 0.1831$_{\pm0.0272}$ & 0.6953$_{\pm0.0723}$ \\
					& SVAE & 0.0212$_{\pm0.0059}$ & 161.0$_{\pm39.33}$ & 0.4315$_{\pm0.1595}$ & 0.00109$_{\pm0.00038}$ & 0.1811$_{\pm0.0378}$ & 0.7223$_{\pm0.0984}$ \\
					& LSTM & 0.0682$_{\pm0.0319}$ & 271.5$_{\pm131.8}$ & -0.8286$_{\pm0.8559}$ & 0.00539$_{\pm0.00357}$ & 0.2784$_{\pm0.0844}$ & -0.3652$_{\pm0.9060}$ \\
					& AOPU & 0.0147$_{\pm0.0003}$ & 177.4$_{\pm4.092}$ & 0.6054$_{\pm0.0094}$ & 0.00076$_{\pm0.00003}$ & 0.1862$_{\pm0.0063}$ & 0.8069$_{\pm0.0092}$ \\
					& RVFLNN & 0.0479$_{\pm0.0555}$ & 106.6$_{\pm74.56}$ & -0.2843$_{\pm1.489}$ & 0.00898$_{\pm0.00208}$ & 0.4238$_{\pm0.3727}$ & -1.2760$_{\pm5.2820}$ \\
					\bottomrule
				\end{tabular}
			\end{threeparttable}
		\end{adjustbox}
	}
\end{table*}

To verify the reliability of the AOPU model's performance and to quantify its comparison with other methods, we implemented two different training strategies. Strategy one involved an early stopping trick and used the best checkpoint to validate the model's performance on the test dataset. Strategy two involved training all models for 40 epochs and using the final checkpoint to test the model performance. All following experiments has batch size set to 64. The outputs of strategy one are presented at Table \ref{tab:comprehensive best experiments}, while the results of strategy two are presented in Table \ref{tab:comprehensive hard experiments}. All NN configurations were subjected to 20 independent repeat experiments, with the mean of the experiments represented by uppercase numbers on the left of the table and the standard deviation by lowercase subscript numbers on the right.

From Table \ref{tab:comprehensive best experiments}, we can intuitively compare the optimal performance among all models. Overall, there is not much difference in final performance among the various networks. 
Notably, almost all MAPE metrics on the Debutanizer dataset exceed 100 due to a sample in the test dataset where the butane content is nearly zero, which significantly distorts the MAPE calculation.
While AOPU performs comparably to other network models in terms of the $\text{R}^2$ metric, its stability is significantly superior, as indicated by much lower standard deviations in the $\text{R}^2$ values compared to all other models.

Further, to more closely align with real industrial application scenarios, if we do not record the optimal checkpoint but instead complete training for 40 epochs as shown in Table \ref{tab:comprehensive hard experiments}, the performance of the models significantly declines. Despite this, AOPU continues to provide stable and reliable performance. In Table \ref{tab:comprehensive hard experiments}, it is noted that using the $\text{R}^2$ metric, AOPU consistently performs best with the sequence length set at 48, and the performance drop from the optimal results calculated using strategy one is minimal. For instance, on the Debutanizer dataset, AOPU's final optimal performance is 0.6054, which is a 14.9\% decrease from the original 0.7118; on the SRU dataset, its best final performance is 0.8069, only a 1.7\% decrease from 0.8209.

Compared to the results on the Debutanizer and SRU datasets, other models show more significant declines: Autoformer dropped from 0.2091 and 0.4366 to -0.2396 and 0.1536 respectively, a decline of 123.9\% and 64.8\%; Informer decreased from 0.5864 and 0.7715 to 0.4047 and 0.6897 respectively, declines of 30.9\% and 10.6\%; DNN dropped from 0.6941 and 0.7944 to 0.5109 and 0.5639 respectively, declines of 26.3\% and 29.0\%; SDAE decreased from 0.7182 and 0.8070 to 0.4470 and 0.7209 respectively, declines of 37.7\% and 10.7\%; SVAE dropped from 0.7180 and 0.8149 to 0.4315 and 0.7397 respectively, declines of 39.9\% and 9.2\%; LSTM went from 0.3097 and 0.7750 to -0.3903 and 0.6280 respectively, declines of 226.0\% and 19.0\%; RVFLNN decreased from 0.7070 and 0.8128 to 0.2844 and 0.5652 respectively, declines of 59.8\% and 30.5\%.

In terms of robustness, AOPU shows a significant improvement. It is observed that AOPU's model performance steadily improves as sequence length increases, in contrast to the other comparison networks, which exhibit large fluctuations. For example, RVFLNN's R$^2$ on the SRU dataset drastically drops from 0.5652 at seq 16 to -0.8785 at seq 32, indicating extreme robustness.

In terms of stability, AOPU not only demonstrates an outstanding advantage in terms of average R$^2$ values but also in standard deviation. From Table \ref{tab:comprehensive hard experiments}, it can be seen that the standard deviation for AOPU between the two training strategies changes only slightly, for instance, from 0.0087 to 0.0094 on the Debutanizer dataset, an increase of about 8.0\%. In contrast, the standard deviation for other networks often increases several-fold, such as Autoformer on the Debutanizer dataset, which increases from an optimal 0.0481 to 0.3287, an increase of about 583.4\%; similar trends are observed with other models.

\subsection{Ablation Study}
\label{ablation study}

In this section, we further investigate the effects of structural designs for augmentation through some ablation studies, examining the impacts of the ReLU piecewise activation function, the Tanh smooth activation function, and normalization on AOPU's performance. It is important to note that if AOPU is trained using direct gradient descent without dual parameter updates, it actually degenerates to an RVFLNN model, and this part of the ablation study has been detailed in section \ref{Comparsion result}.

From Table \ref{tab:acti and norm influence} we can draw two conclusions: The first is normalization significantly impairs AOPU's model performance. The second it ReLU piecewise non-linear activation function suits worse for AOPU than the Tanh activation function. As previously analyzed in \ref{Network's mechanism} where both the input data $\tilde{x}$ and $y$ should to be zero mean, hence reducing the covariance operator $\textup{R}$ to an inner product operator. However, piecewise linear functions like ReLU and LeakyReLU are not zero-mean, which violates such assumptions.

\begin{table*}
	\caption{AOPU evaluation metrics on the Debutanizer and SRU datasets under various combinations of activation functions and layer normalization settings.}
	\label{tab:acti and norm influence}
	\centering
	{\scriptsize
		\setlength{\tabcolsep}{4pt}
		\begin{adjustbox}{center}
			\begin{threeparttable}
				\begin{tabular}{cc|c|ccc|ccc}
					\toprule
					\multicolumn{3}{c}{Structure} & \multicolumn{6}{c}{Dataset \& Metric} \\
					\midrule
					\multicolumn{2}{c}{Acti} & {Norm} & \multicolumn{3}{c|}{Debutanizer} & \multicolumn{3}{c}{SRU} \\
					Relu & Tanh & LaNorm & MSE & MAPE & R$^2$ & MSE & MAPE & R$^2$ \\
					\midrule

					\CheckedBox & $\Box$ & $\Box$ & 0.0162$_{\pm0.00030}$ & 183.2$_{\pm4.8862}$ & 0.5654$_{\pm0.0082}$ & 0.00074$_{\pm0.00002}$ & 0.1838$_{\pm0.0031}$ & 0.8115$_{\pm0.0060}$ \\
					
					$\Box$ & \CheckedBox & $\Box$ & 0.0103$_{\pm0.00021}$ & 148.6$_{\pm6.3927}$ & 0.7216$_{\pm0.0059}$ & 0.00077$_{\pm0.00003}$ & 0.1902$_{\pm0.0049}$ & 0.8026$_{\pm0.0084}$ \\

					\CheckedBox & $\Box$ & \CheckedBox & 0.0633$_{\pm0.02113}$ & 117.2$_{\pm57.219}$ & -0.6961$_{\pm0.5650}$ & 0.58831$_{\pm0.17940}$ & 4.5829$_{\pm0.6003}$ & -147.98$_{\pm45.422}$ \\
					
					$\Box$ & \CheckedBox & \CheckedBox & 0.0189$_{\pm0.00662}$ & 157.6$_{\pm62.892}$ & 0.4930$_{\pm0.1773}$ & 0.00205$_{\pm0.00082}$ & 0.3149$_{\pm0.0619}$ & 0.4803$_{\pm0.2055}$ \\
					
					\bottomrule
				\end{tabular}
			\end{threeparttable}
		\end{adjustbox}
	}
\end{table*}

\begin{table*}
	\caption{Comprehensive experiments on activation function influence to AOPU}
	\label{tab:acti and norm influence2}
	\centering
	{\scriptsize
		\setlength{\tabcolsep}{4pt}
		\begin{adjustbox}{center}
			\begin{threeparttable}
				\begin{tabular}{c|c|ccc|ccc}
					\toprule
					\multicolumn{2}{c}{Structure} & \multicolumn{6}{c}{Dataset \& Metric} \\
					\midrule
					\multirow{2}{*}{Class} & \multirow{2}{*}{Acti} & \multicolumn{3}{c|}{Debutanizer} & \multicolumn{3}{c}{SRU} \\
					& & MSE & MAPE & R$^2$ & MSE & MAPE & R$^2$ \\
					\midrule
					\multirow{5}{*}{zero-mean}
					& Hard Shrink & 0.0103$_{\pm0.0003}$ & 151.9$_{\pm6.359}$ & 0.7236$_{\pm0.0081}$ & 0.00076$_{\pm0.00002}$ & 0.1882$_{\pm0.0037}$ & 0.8069$_{\pm0.0059}$ \\
					& Tanh & 0.0104$_{\pm0.0004}$ & 152.0$_{\pm6.425}$ & 0.7202$_{\pm0.0114}$ & 0.00076$_{\pm0.00004}$ & 0.1894$_{\pm0.0052}$ & 0.8056$_{\pm0.0106}$ \\
					& Tanh Shrink & 0.0103$_{\pm0.0002}$ & 152.7$_{\pm5.805}$ & 0.7221$_{\pm0.0068}$ & 0.00077$_{\pm0.00003}$ & 0.1891$_{\pm0.0056}$ & 0.8043$_{\pm0.0093}$ \\
					& Soft Sign & 0.0103$_{\pm0.0003}$ & 152.8$_{\pm6.309}$ & 0.7225$_{\pm0.0091}$ & 0.00077$_{\pm0.00002}$ & 0.1905$_{\pm0.0033}$ & 0.8040$_{\pm0.0068}$ \\
					& Soft Shrink & 0.0104$_{\pm0.0002}$ & 152.6$_{\pm4.548}$ & 0.7201$_{\pm0.0060}$ & 0.00076$_{\pm0.00003}$ & 0.1872$_{\pm0.0050}$ & 0.8074$_{\pm0.0081}$ \\
					\noalign{\vskip 3pt}
					\noalign{\vskip -2pt} \cline{1-8}
					\noalign{\vskip 3pt}
					\multirow{5}{*}{non-zero-mean} 
					& Sigmoid & 0.0157$_{\pm0.0004}$ & 192.3$_{\pm6.336}$ & 0.5777$_{\pm0.0124}$ & 0.00080$_{\pm0.00003}$ & 0.1925$_{\pm0.0048}$ & 0.7955$_{\pm0.0089}$ \\
					& Relu6 & 0.0171$_{\pm0.0004}$ & 187.9$_{\pm5.502}$ & 0.5416$_{\pm0.0109}$ & 0.00074$_{\pm0.00003}$ & 0.1836$_{\pm0.0051}$ & 0.8111$_{\pm0.0076}$ \\
					& RRelu & 0.0165$_{\pm0.0002}$ & 186.3$_{\pm5.224}$ & 0.5582$_{\pm0.0074}$ & 0.00076$_{\pm0.00004}$ & 0.1858$_{\pm0.0047}$ & 0.8072$_{\pm0.0105}$ \\
					& Hard Swish & 0.0103$_{\pm0.0002}$ & 149.2$_{\pm5.500}$ & 0.7232$_{\pm0.0078}$ & 0.00077$_{\pm0.00003}$ & 0.1900$_{\pm0.0036}$ & 0.8026$_{\pm0.0076}$ \\
					& Mish & 0.0103$_{\pm0.0003}$ & 151.4$_{\pm6.260}$ & 0.7229$_{\pm0.0098}$ & 0.00079$_{\pm0.00003}$ & 0.1927$_{\pm0.0054}$ & 0.7999$_{\pm0.0089}$ \\
					\bottomrule
				\end{tabular}
			\end{threeparttable}
		\end{adjustbox}
	}
\end{table*}

To validate the analysis regarding the effects of zero-mean and non-zero-mean activation functions on AOPU's performance, an additional comparative experiment was conducted. This experiment included 20 independent repetitions for activation functions classified into zero-mean, Hard Shrink, Tanh, Tanh Shrink, Soft Sign, and Soft Shrink, and non-zero-mean, Sigmoid, Relu6, RRelu, Hardswish, and Mish. The results are listed in Table \ref{tab:acti and norm influence2}

The experimental results largely confirmed the hypotheses outlined previously. In the zero-mean group, whether on the Debutanizer or SRU dataset, fluctuations in MSE, MAPE, and $\text{R}^2$ metrics were consistently controlled within 1\% (with the maximum $\text{R}^2$ fluctuation being 0.48\%, from 0.7236 to 0.7201). Conversely, in the non-zero-mean group, Sigmoid, Relu6, and RRelu all demonstrated notable performance declines on the Debutanizer dataset. Notably, although Hard Swish and Mish are classified as non-zero-mean activation functions, they did not negatively impact AOPU's performance. This could likely be attributed to the fact that Hard Swish and Mish are approximately zero-mean near the zero index, unlike Sigmoid, Relu6, and RRelu, which are non-zero-mean across any arbitrary small neighborhoods.

\section{Conclusion and Limitation}
This paper introduces a novel NN regression model, AOPU, which is grounded in solid mathematics basis and validated through extensive experiments. The results demonstrate its superior performance, robustness, and training stability. The development of AOPU lays the foundation for the practical implementation of deep learning soft sensor techniques in industrial processes and provides guidance for subsequent control, monitoring, and optimization management of these processes.
The introduction of RR also illuminates a promising and valuable direction for exploring the design of augmentation models. 
Such prospective topics of value encompass how to reduce the sensitivity of AOPU to batch size and sequence length, how to derive the NG optimization of the augmentation model, and how to bolster the nonlinear modeling capability of the augmentation model. 

We note that AOPU is not a "plug-and-play" model; it requires adjustments based on actual data conditions. AOPU necessitates a clear understanding of the RR distribution of data intended for application to guide the selection of batch size and sequence length hyperparameters. This requirement stems from the inherent matrix inversion operations in AOPU. When the RR value is too low, noise during the AOPU training process can greatly exceed the effective information, potentially leading to model divergence as Appendix \ref{supp figure} discusses.

\newpage

\bibliographystyle{nips}
\bibliography{references}

\newpage
\appendix

\section{Network's mechanism}
\label{Network's mechanism}

In this section, we first demonstrate that AOPU implements a MVE through NN, explain the relationship between data augmentation, minimum variance, and orthogonal projection. We then discuss the physical significance and necessity of the dual parameter from NGD perspective.

\subsection{From MVE perspective}
\label{sec:MVE}

We start by giving the definition of the MVE,
\begin{definition}
	Given the independent variable $x$ and the dependent variable $y$, $f^{*}(y|x)$ is said to be the MVE for $y$ if the following hold,
	\begin{equation}
		\begin{aligned}
			\textup{E}[(y-f^*(y|x))^2] \leq \textup{E}[(y-f(y|x))^2]
		\end{aligned}
	\end{equation}
	where $\textup{E}[\cdot]$ denotes the expectation operator, and $f(y|x)$ represents any unbiased arbitrary estimation function for $y$ given $x$.
	\label{def:1}
\end{definition}

According to definition \ref{def:1}, it is straightforward that the MVE is the optimal unbiased estimator under the Mean Squared Error loss metric, providing a performance boundary for all regression networks. However, the solution to the MVE, detailed in Appendix \ref{g-mve}, represented as $\int yp(y|x)dy$ where $p(\cdot)$ denotes the probability operator, is challenging to determine. Since the prior knowledge of the likelihood distribution $p(y|x)$ is not accessible, this integral is difficult to solve. Instead of delving into modeling the likelihood, AOPU turns to referencing solvable linear MVE operators for network design. Specifically, when the function form of $f(y|x)$ is constrained to be linear with respect to $x$, it can be set as $f(y|x)=W_{mve}x+b$, with the solution $W_{mve}=\textup{R}_{yx}\textup{R}_{xx}^{-1}$ and $b=\textup{E}[y]-W_{mve}\textup{E}[x]$, where $\textup{R}_{ab}$ represents the covariance matrix $\textup{E}\left [(a-\textup{E}[a])(b-\textup{E}[b])^T\right ]$. For clarity, the proof is listed in Appendix \ref{l-mve}.

Given that variables $x$ and $y$ have been normalized to have zero mean, i.e., $\textup{E}[x]=0$ and $\textup{E}[y]=0$, it follows that $b=0$, and the covariance operator $\textup{R}$ degenerates into an inner product operation. Revisiting the loss function of AOPU, it is evident from Eq. \ref{loss func} that AOPU essentially estimates the parameters $W_{mve}$ of the linear MVE. Here the $(\tilde{x}^T\tilde{x})^{-1}$ is aligned with the $\textup{R}_{\tilde{x}\tilde{x}}^{-1}$, and $\tilde{x}^TD(\tilde{W})$ ought correspond to the estimation of the input-output covariance matrix. It is noteworthy that the parameter $W_{mve}$ is not the estimator's output. Therefore, by approximating $W_{mve}$ in the loss function, AOPU implies an important assumption: unlike other regression algorithms that explicitly model the mapping relationship from $\tilde{x}$ to $y$, i.e., $\tilde{x} \rightarrow y$, AOPU implicitly models the relationship $\tilde{x} \rightarrow \tilde{x}y$.  Given that $\tilde{x} \in \mathbb{R}^{d+h,b}$ and $y \in \mathbb{R}^{b,1}$, the key to deriving $y$ from known $\tilde{x}$ and $\tilde{x}y$ lies in the requirement that $\tilde{x}$ must be column-full-rank. This requirement aligns with the numerical stability needs during the computation process of AOPU, thereby establishing a self-consistent mathematical framework for the unit.

The geometric interpretation of the linear minimum variance estimator as orthogonal projection underpins the naming of the AOPU. AOPU differs from the orthogonal projection in threefold: (1) Orthogonal projection is a non-parametric batch algorithm, whereas AOPU operates as a parametric, gradient-based mini-batch optimization algorithm. (2) Orthogonal projection requires the covariance matrix's inverse to exist definitively, whereas AOPU can employ approximate inverses for its computations. (3) Orthogonal projection strictly adheres to linear minimum variance estimation, but AOPU introduces non-linearity through data augmentation, allowing it to serve as a versatile minimum variance estimator. The data augmentation techniques illustrated in Fig. \ref{track and untrack} are critical for enhancing the expressive capabilities of AOPU. We initially improve model expressiveness through a fixed, randomly initialized Gaussian matrix $\hat{G}$. However, this approach remains confined within the linear transformation. Consequently, in subsequent experiments, the inputs are further augmented using $\text{LeakyRelu}(\hat{G}^Tx)$, pushing the model beyond linear transformations.

\subsection{From NGD perspective}

In this section, we introduce NGD, including the computation of FIM, how to reduce the computational complexity of NGD through EM, and ultimately demonstrate that the truncated gradient of AOPU is an approximated NG.

We begin with an introduction to the most basic optimization algorithm used in neural network training, Gradient Descent (GD). Assuming the network parameters are represented as $\lambda$ ($\lambda$ represents $\tilde{W}$ in AOPU), and $\nabla$ denotes the gradient, GD can be defined by the following equation:
\begin{equation}
	\begin{aligned}
		\text{GD: \quad} \lambda_{t+1}=\lambda_{t}-\alpha\nabla_{\lambda_{t}}\mathcal{L}(\lambda_{t})
	\end{aligned}
	\label{GD}
\end{equation}
where $\alpha > 0$ represents the network's learning rate. Many optimization algorithms assist the network in escaping local optima and accelerating convergence by incorporating momentum gradients and adaptive learning rates. However, fundamentally, these are first-order optimization methods (considering only first-order derivatives) and typically exhibit suboptimal performance in practice. NG, by calculating the FIM, exploits the information geometry of the model's output distribution to accelerate convergence more effectively. NGD can be defined by the following equation:
\begin{equation}
	\begin{aligned}
		\text{NGD: \quad} \lambda_{t+1}= \lambda_{t} - \beta F(\lambda_{t})^{-1}  \nabla_{\lambda_{t}} \mathcal{L} (\lambda_{t})
	\end{aligned}
\end{equation}
where $\beta$, often set to 1, is a scale factor. $F(\lambda_{t})$ is the negative expectation of the second-order derivative regarding the network parameter $\lambda_{t}$, expressed as $F(\lambda_{t})=-\textup{E}_{p(\hat{y})}[\nabla^2_{\lambda_{t}}\log{p(\hat{y}|\lambda_{t})}]$. The essence of the FIM is to act as a preconditioning matrix that properly scales the gradient on the manifold of the parameter space. The primary difference between NGD and GD is that FIM accounts for dependencies among parameters, unlike the assumption of independent gradients for each parameter as in Eq. \ref{GD} with $\alpha$. NGD generally exhibits better performance compared to GD when FIM is well-defined.

The computational cost of calculating the FIM and its inverse is substantial, i.e., cubic in terms of time complexity and quadratic in terms of space complexity. However, for specific model structures, the computation of NGD can be as straightforward as that of GD. We provide the definition of the exponential family as follows,
\begin{equation}
	\begin{aligned}
		p(z|\lambda):=h(z)\exp{[\Braket{\phi(z),\lambda}-A(\lambda)]}
	\end{aligned}
\end{equation}
where $p(z|\lambda)$ is defined as an exponential family function with the structure given on the right-hand-side, $h(z)$ is the base measure, $\phi(z)$ is the sufficient statistics, $\lambda$ is the natural parameter,  and $A(\lambda)$ is the log-partition function. The expectation-parameter is defined as the expected value of the sufficient statistics, expressed as $m(\lambda)=\textup{E}_{p(z|\lambda)}[\phi(z)]$. For exponential family distributions, we propose the following proposition,
\begin{proposition}
	The FIM of $p(z|\lambda)$ with respect to $\lambda$ is equivalent to the gradient of the expectation parameter $m(\lambda)$ with respect to $\lambda$.
	\label{pro:2}
\end{proposition}
\begin{proof*}
	In Appendix \ref{proof:pro2}.
\end{proof*}
Therefore, the NGD can be simply expressed as follows:
\begin{equation}
	\begin{aligned}
		\text{NGD: \quad} \lambda_{t+1}=\lambda_{t}-\beta\nabla_{m(\lambda_{t})} \mathcal{L}(\lambda_{t})
	\end{aligned}
\end{equation}

By utilizing the MSE loss function, we can establish a connection between AOPU and NGD. From Eq. \ref{loss func}, it is known that the output of AOPU is the inner product of the model parameter $\tilde{W}$ and the augmented variable $\tilde{x}$. Supposed the output is viewd as an Gaussian distribution characterized by mean $\tilde{x}^T\tilde{W}$ and covariance $I$, where $I$ is an appropriately dimensioned identity matrix. The following optimization objective can be proven to be equivalent to MSE:
\begin{equation}
	\begin{aligned}
		\hat{\mathcal{L}}(\tilde{W})=&\textup{E}_{\mathcal{N}(\hat{y}|\tilde{x}^T\tilde{W},I)}[(y-\hat{y})^T(y-\hat{y})] \\
		=&(y-\hat{y})^T(y-\hat{y})+\textup{Tr}(I) \\
		\approxeq&\text{MSE}(y,\hat{y})
	\end{aligned}
\end{equation}
where $\approxeq$ denotes the equivalence up to constants.

\begin{proposition}
	The NG of $\hat{\mathcal{L}}(\tilde{W})$ with respect to $\tilde{W}$ is approximately equivalent to the gradient of $\hat{\mathcal{L}}(D(\tilde{W}))$ with respect to dual parameter $D(\tilde{W})$.
	\label{pro:3}
\end{proposition}
\begin{proof*}
	In Appendix \ref{proof:pro3}
\end{proof*}
Proposition \ref{pro:3} valids the effectiveness of using truncated gradients for parameter updates in AOPU. 
Fact that $\hat{\mathcal{L}}(D(\tilde{W}))$ is actuallt the aforementioned loss function $\mathcal{L}$, and it can only be an approximation of $\hat{\mathcal{L}}(\tilde{W})$ because the inverse of $\tilde{x}^T\tilde{x}$ may not be well-defined as discussed in sections \ref{sec:network structure} and \ref{sec:MVE}. Therefore, RR not only serves as a measure of AOPU's numerical stability and its approximation degree to MVE but is also employed to quantify how closely truncated gradient approximates the NG.

\section{Convergence Analysis}
\label{convergence analysis}
In this section, we are going to analyze the convergence of AOPU referencing the conclusions from \cite{Stochastic-Mirror-Descent,On-the-Convergence-of-Mirror-Descent}. 
We eventually demonstrate that under the condition $\tilde{x}$ is column-full-rank, AOPU converges to the optimal solution almost surely. 
Firstly, thanks to the trackability of parameters, AOPU is capable of being proven a coherent optimization problem under a strict assumption, which agrees on previous analysis, is made about the distribution of the observed samples $y$ during the proof; AOPU's truncated gradient is then proven to structurally ensure consistency with the Stochastic Mirror Descent (SMD), specifically that the dual parameters correspond directly to the mirror map; the assumptions in \cite{Stochastic-Mirror-Descent} about regularity (assumption 3), differentiability (assumption 1), and bounded second moments with Lipschitz continuity (assumption 2) are also proven to be met under the condition $\tilde{x}$ is column-full-rank. Finally, by referencing theorem 3.4 from \cite{Stochastic-Mirror-Descent}, we prove that AOPU can enter arbitrarily small neighborhood of the optimal parameter solution $\mathcal{\tilde{W}}^*$.

\begin{definition}
	The optimization problem $\min{\tilde{\mathcal{L}}(W)}$ is said to be coherent if
	\begin{equation}
		\label{eq:coherent}
		\begin{aligned}
			\Braket{\nabla \textup{E}[\tilde{\mathcal{L}}(W)], W-W^*} \ge 0\text{\quad for all }W \in \mathcal{W},W^* \in \mathcal{W}^*
		\end{aligned}
	\end{equation}
	with equality holds if and only if $W \in \mathcal{W}^*$, where $\mathcal{W}$ is the feasible parameter space and $\mathcal{W}^*$ is additionally constrainted by $\mathcal{W}^*=\arg\min \tilde{\mathcal{L}}$.
	\label{def:2}
\end{definition}

\begin{proposition}
	\label{pro:4}
	If the observed sample $y$ is characterized by an underlying $D(\tilde{W})^* \in \mathcal{D(\tilde{W})}^*$, $D$ for short, AOPU's training objective is coherent with respect to $D$.
\end{proposition}
\begin{proof*}
	In Appendix \ref{proof:pro4}
\end{proof*}

Referring the work of \cite{Stochastic-Mirror-Descent,Optimistic-Dual-Extrapolation}, Definition \ref{def:2} introduces the concept \textbf{coherent} which involves the analysis of the first-order derivatives of the loss function. For conventional NN, due to the untrackability of parameters, analyzing parameter gradients is exceedingly challenging. Although some studies \cite{RN47,Can-We-Remove,Optimizing-Neural-Networks-with-Kronecker-factored} have explored the parameters' local characteristics, analyzing their global properties remains difficult. Proposition \ref{pro:4} summarizes the properties of coherence of AOPU briefly.

\begin{proposition}
	\label{pro:5}
	If the regularizer in $Q$ is characterized by square Mahalanobis distance with covariance matrix $\Sigma=\tilde{x}\tilde{x}^T$ instead of Euclidean distance $\Sigma=I$, AOPU's training strategy is identical to SMD.
\end{proposition}
\begin{proof*}
	In Appendix \ref{proof:pro5}
\end{proof*}

In SMD, each iteration involves calculating the stochastic gradient from the model's current state, updating within the dual space, and then mapping back to the parameter space. This process is outlined in Algorithm \ref{alg:smd}, where $X$ and $Y$ represent the parameters and dual parameters, respectively, and subscripts denote the iteration number. 
$Q$ represents the mirror map, which is defined as follows,
\begin{equation}
	Q(Y)=\arg\max_{X}\Braket{X,Y}-h(X)
\end{equation}
where $h$ acts as a regularizer. Proposition \ref{pro:5} summarizes the connection between AOPU's training and SMD by carefully selecting regularizer.
\begin{algorithm}
	\caption{Stochastic mirror descent}
	\label{alg:smd}
	\begin{algorithmic}[1]
		\REQUIRE Initial $Y_0$
		\STATE $n \gets 0$
		\REPEAT
		\STATE $X_n = Q(Y_n)$
		\STATE $Y_{n+1} = Y_n - \alpha_{n+1} \nabla \tilde{\mathcal{L}}(X_n)$
		\STATE $n \gets n + 1$
		\UNTIL{end}
		\RETURN solution candidate $X_n$
	\end{algorithmic}
\end{algorithm}

The assumption about regularity fundamentally guarantees the continuity of the Fenchel coupling at the point where $Y$ equals the subgradient of $h$ with respect to $X$, and can typically be considered trivially satisfied.
Regarding the properties of differentiability and bounded second moments and Lipschitz continuity, the objective function of AOPU is typically quadratic, thus both its first and second derivatives exist and are linearly related to $(\tilde{x}^T\tilde{x})^{-1}$. These assumptions are only satisfied when $\tilde{x}$ is column-full-rank, i.e., when the inverse is well-defined.

Combining the conclusions from the proofs discussed above with theorem 3.4 from \cite{Stochastic-Mirror-Descent}, it can be determined that AOPU's dual parameters $D(\tilde{W})$ always converge to the optimal values under the constraints. 
Essentially, the primary difference between using $\tilde{W}$ for inference and $D(\tilde{W})$ for training is term $(\tilde{x}^T\tilde{x})^{-1}$.
Therefore, when $\tilde{x}$ is column-full-rank, the convergence of the dual parameters $D(\tilde{W})$ is equivalent to the convergence of the parameters $\tilde{W}$. This conclusion is consistent with the results from the earlier MVE analysis and highlights the important role of RR.

The proof of convergence complements the final piece of the puzzle for the deployment of AOPU in advanced industrial applications. It provides solid theoretical support in the aspects of derivation processes, optimization procedures, state monitoring, and convergence assurance. It can be confidently stated that AOPU is ready to be applied in industrial soft sensing, having established robust foundations for operational reliability and efficacy.

\section{Mathematic Proof}
\label{mathematic proof}
\subsection{Solution to General Minimum Variance Estimator}
\label{g-mve}
In this subsection we are about to prove that given $x$, the solution to the general minimum variance estimator of $y$ is $\int yp(y|x)dy$, i.e., $\textup{E}_{y|x}[y]$. Since it is the expectation of likelihood, this result is intuitive to prove. Rewrite the covariance calculation in the following,
\begin{equation}
	\begin{aligned}
		&\textup{E}\left[ (y-f(y|x))(y-f(y|x))^T\right] \\
		=&\textup{E}\left[ (y+\textup{E}_{y|x}[y]-\textup{E}_{y|x}[y]-f(y|x))(y+\textup{E}_{y|x}[y]-\textup{E}_{y|x}[y]-f(y|x))^T\right]\\
		=& \textup{E}\left[ (y-\textup{E}_{y|x}[y])(y-\textup{E}_{y|x}[y])^T\right] + \textup{E}\left[ (\textup{E}_{y|x}[y]-f(y|x))(\textup{E}_{y|x}[y]-f(y|x))^T\right]+\\
		&\textup{E}\left[ (y-\textup{E}_{y|x}[y])(\textup{E}_{y|x}[y]-f(y|x))^T\right]+\textup{E}\left[ (\textup{E}_{y|x}[y]-f(y|x))(y-\textup{E}_{y|x}[y])^T\right]
	\end{aligned}
\end{equation}

Noting that $f(y|x)$ is not a conditional probabilistic distribution representation, it denotes a function that takes $x$ as input, and the output of such function is regarded as an estimator of $y$. In conclusion, $f(y|x)$ is fundamentally independent of y, therefore, for the term $\textup{E}\left[ (\textup{E}_{y|x}[y]-f(y|x))(y-\textup{E}_{y|x}[y])^T \right]$ we can rewrite it into,
\begin{equation}
	\begin{aligned}
		&E\left[ (\textup{E}_{y|x}[y]-f(y|x))(y-\textup{E}_{y|x}[y])^T \right] \\
		=&\int\int (\textup{E}_{y|x}[y]-f(y|x))(y-\textup{E}_{y|x}[y])^T p(x,y)dxdy \\
		=&\int (\textup{E}_{y|x}[y]-f(y|x)) \left( \int (y-\textup{E}_{y|x}[y])^Tp(y|x)dy \right) p(x)dx\\
		=&\int (\textup{E}_{y|x}[y]-f(y|x))  (\textup{E}_{y|x}[y]-\textup{E}_{y|x}[y])^T p(x)dx\\
		=&0
	\end{aligned}
	\label{eq:zero1}
\end{equation}

The conclusion also applies to the term $\textup{E}\left[ (y-\textup{E}_{y|x}[y])(\textup{E}_{y|x}[y]-f(y|x))^T \right]$. Consequently, the last two terms in Eq. \ref{eq:zero1} consistently equal zero. Given that $\textup{E}\left[ (\textup{E}_{y|x}[y]-f(y|x))(\textup{E}_{y|x}[y]-f(y|x))^T \right]$ is semi-positive definite, it follows that $\textup{E}\left[ (y-\textup{E}_{y|x}[y])(y-\textup{E}_{y|x}[y])^T \right]$ establishes a lower bound for $\textup{E}\left[ (y-f(y|x))(y-f(y|x))^T \right]$. Equality holds if and only if  $f(y|x)=\textup{E}_{y|x}[y]$ which completes the proof.

\subsection{Solution to Linear Minimum Variance Estimator}
\label{l-mve}
In this subsection, we are about to prove that the solution to the linear minimum variance estimator is $W_{mve}=\textup{R}_{yx}\textup{R}_{xx}^{-1}$ and $b=E[y]-W_{mve}\textup{E}[x]$. Initially, it is straightforward to see that the value of $b$ renders the estimator unbiased. By simply taking the expectation, we can complete the proof. Again we rewrite the covariance calculation in the following,
\begin{equation}
	\begin{aligned}
		&\textup{E}\left[ (y-f(y|x))(y-f(y|x))^T\right] \\
		=&\textup{E}\left[ (y-\textup{E}[y]+W_{mve}\textup{E}[x]-W_{mve}x)(y-\textup{E}[y]+W_{mve}\textup{E}[x]-W_{mve}x)^T\right]
	\end{aligned}
	\label{eq:covariance transformation}
\end{equation}

Incorporating the covariance matrices $\textup{R}_{yy}=\textup{E}\left[ (y-\textup{E}[y])(y-\textup{E}[y])^T \right]$, $\textup{R}_{xx}=\textup{E}\left[ (x-\textup{E}[x])(x-\textup{E}[x])^T \right]$, and $\textup{R}_{xy}=\textup{E}\left[ (x-\textup{E}[x])(y-\textup{E}[y])^T \right]$. We can reformulate Eq. \ref{eq:covariance transformation} as $\textup{R}_{yy}+W_{mve}\textup{R}_{xx}W_{mve}^T-\textup{R}_{yx}W_{mve}^T-W_{mve}\textup{R}_{xy}$. Upon simplification, this equation transforms to,
\begin{equation}
	\begin{aligned}
		&\textup{R}_{yy}+W_{mve}\textup{R}_{xx}W_{mve}^T-\textup{R}_{yx}W_{mve}^T-W_{mve}\textup{R}_{xy} \\
		=&W_{mve}\textup{R}_{xx}W_{mve}^T-W_{mve}\textup{R}_{xx}\textup{R}_{xx}^{-1}\textup{R}_{xy}-\textup{R}_{yx}\textup{R}_{xx}^{-1}\textup{R}_{xx}W_{mve}^T+\textup{R}_{yy} + \textup{R}_{yx}\textup{R}_{xx}^{-1}\textup{R}_{xy} - \textup{R}_{yx}\textup{R}_{xx}^{-1}\textup{R}_{xy}\\
		=&(W_{mve}-\textup{R}_{yx}\textup{R}_{xx}^{-1})\textup{R}_{xx}(W_{mve}-\textup{R}_{yx}\textup{R}_{xx}^{-1})^T+\textup{R}_{yy}-\textup{R}_{yx}\textup{R}_{xx}^{-1}\textup{R}_{xy}
	\end{aligned}
\end{equation}

Noting that $(W_{mve}-\textup{R}_{yx}\textup{R}_{xx}^{-1})\textup{R}_{xx}(W_{mve}-\textup{R}_{yx}\textup{R}_{xx}^{-1})^T$ is again semi-positive definite, indicating that the optimal $W_{mve}$ is identical to $\textup{R}_{yx}\textup{R}_{xx}^{-1}$, which completes the proof.

\subsection{Proof to Proposition \ref{pro:1}}
\label{sec:proof1}
Suppose there exists an operator $T$ independent of $x$ such that for a given $W$ and any inputs $x_1$ and $x_2$, the Eq. \ref{eq-propos1} holds. From the linearity property of operators, it follows that,
\begin{equation}
	\begin{aligned}
		\textup{acti}(Wx_1)+\textup{acti}(Wx_2)=T(W)(x_1+x_2)\\
		\textup{acti}(Wx_1)+\textup{acti}(Wx_2)=\textup{acti}(W(x_1+x_2))
	\end{aligned}
	\label{eq-proof1}
\end{equation}
Due to the nonlinearity of the activation function, it is clear that Eq. \ref{eq-proof1} doesn't hold, consequently completing the proof.

\subsection{Proof to Proposition \ref{pro:2}}
\label{proof:pro2}
We are about to give concise and precise proof in this section, starting by proving the connection between the expectation-parameter and the log-partition function.
\begin{proposition}
	The expectation-parameter equals to the gradient of log-partition function with respect to natural parameter.
	
\end{proposition}
\begin{proof*}
	Since EM represents a probability distribution, the log-partition function acts as a normalizing factor, thus the following identity holds true,
	\begin{equation}
		A(\lambda)=\log \int{h(z)\exp{(\Braket{\phi(z),\lambda})}dz}.
	\end{equation}
	
	Therefore, expectation-parameter could be derived from differentiating $A(\lambda)$ with respect to $\lambda$.
	
	\begin{equation}
		\begin{aligned}
			\nabla_{\lambda}A(\lambda) &= \nabla_{\lambda} \log \int{h(z)\exp{(\Braket{\phi(z),\lambda})}dz} \\
			&= \frac{\nabla_{\lambda}\int{h(z)\exp{(\Braket{\phi(z),\lambda})}dz}}{\int{h(z)\exp{(\Braket{\phi(z),\lambda})}dz}} \\
			&=\frac{\nabla_{\lambda}\Braket{\phi(z),\lambda} 
				\left( \int{h(z)\exp{(\Braket{\phi(z),\lambda})}dz} \right)}{\int{h(z)\exp{(\Braket{\phi(z),\lambda})}dz}} \\
			&=\textup{E}_{p(z|\lambda)}{\phi(z)}
		\end{aligned}
	\end{equation}
\end{proof*}

Clearly, $A(\lambda)$ is the only term that is second-order derivable in the score function $\log{p(z|\lambda)}$. The FIM can then be intuitively derived from its definition.
\begin{equation}
	\begin{aligned}
		F(\lambda)&=-\textup{E}_{p(z|\lambda)}[\nabla^2_{\lambda}\log{p(z|\lambda)}] \\
		&= -\textup{E}_{p(z|\lambda)}[-\nabla^{2}_{\lambda}A(\lambda)]\\
		&= \nabla m(\lambda)
	\end{aligned}
\end{equation}

\subsection{Proof to Proposition \ref{pro:3}}
\label{proof:pro3}
We first reiterate that treating the output $g(\hat{y}|x)$ as a Gaussian distribution is merely a prior assumption and does not alter the structure or computation of AOPU. Representing this Gaussian distribution as the minimal EM can be expressed as follows,
\begin{equation}
	\begin{aligned}
		\mathcal{N}(\hat{y}|\tilde{x}^T\tilde{W},I) =& (2\pi)^{-\frac{d+h}{2}} \left| I \right|^{-\frac{1}{2}}\exp \left[ -\frac{1}{2} \Braket{\hat{y}-\tilde{x}^T\tilde{W},\hat{y}-\tilde{x}^T\tilde{W}} \right] \\
		=& (2\pi)^{-\frac{d+h}{2}} \left| I \right|^{-\frac{1}{2}} \exp \left[ \Braket{\tilde{x}\hat{y}, \tilde{W}} -\frac{1}{2}\left( \hat{y}^T\hat{y}+\tilde{W}^T\tilde{x}\tilde{x}^T\tilde{W} \right)  \right] \\
	\end{aligned}
\end{equation}
Under this representation, the sufficient statistics and the natural parameter are respectively $\tilde{x}\hat{y}$ and $\tilde{W}$. From this, by the definition of the expectation parameter, we can calculate $m(\tilde{W})=\textup{E}_{\mathcal{N}(\hat{y}|\tilde{x}^T\tilde{W},I)}[\tilde{x}\hat{y}]$ equals to $\tilde{x}\tilde{x}^T\tilde{W}$ which is exactly identical to $D(\tilde{W})$. According to proposition \ref{pro:2}, the FIM with respect to $\tilde{W}$ is equivalent to the gradient of $D(\tilde{W})$ with respect to $\tilde{W}$. Thus, by introducing $D(\tilde{W})$, we can accelerate the NGD computation with respect to $\tilde{W}$ as shown below,

Note that $\hat{\mathcal{L}}(\tilde{W})$ is equivalent to MSE. To compute the gradient of $\hat{\mathcal{L}}$ at $D(\tilde{W})$ using automatic differentiation tools and avoid complex algebraic operations, we design $\hat{\mathcal{L}}(D(\tilde{W}))$ as,
\begin{equation}
	\begin{aligned}
		\hat{\mathcal{L}}(D(\tilde{W}))=&\textup{E}_{\mathcal{N}(\hat{y}|(\tilde{x}^T\tilde{x})^{-1}\tilde{x}^TD(\tilde{W}),I)}[(y-\hat{y})^T(y-\hat{y})] \\
	\end{aligned}
\end{equation}
Clearly, $\hat{\mathcal{L}}(D(\tilde{W}))$ is identical to the $\mathcal{L}$ introduced in section \ref{sec:network structure}.

\subsection{Proof to Proposition \ref{pro:4}}
\label{proof:pro4}

We now assume the observed sample $y$ is fully characterized by dual parameter $D$ addition with a zero-mean random variable $\epsilon$. Such constraint implies that there exists an optimal parameter set $\mathcal{D}^*$ which fully captures the mean trend of $y$, i.e., $y=(\tilde{x}^T\tilde{x})^{-1}\tilde{x}^T\mathcal{D}^*+\epsilon$. The coherence definition could be rewritten as follows,
\begin{equation}
	\label{eq:conherence-proof}
	\begin{aligned}
		&\Braket{\nabla \textup{E}[\hat{\mathcal{L}}(D)], D-D^*} \\
		=&\textup{E}\left[ \Braket{\nabla\hat{\mathcal{L}}(D), D-D^*} \right] \\
		=&\textup{E}\left[ \Braket{(\tilde{x}^T\tilde{x})^{-1}\tilde{x}^TD-y, (\tilde{x}^T\tilde{x})^{-1}\tilde{x}^T(D-D^*)} \right] \\
		=&\textup{E}\left[ \Braket{(\tilde{x}^T\tilde{x})^{-1}\tilde{x}^TD-y, (\tilde{x}^T\tilde{x})^{-1}\tilde{x}^T(D-D^*)-y+y} \right] \\
		=&\textup{E}\left[ \hat{\mathcal{L}}(D) \right] + \textup{E} \left[\Braket{(\tilde{x}^T\tilde{x})^{-1}\tilde{x}^TD-y, y-(\tilde{x}^T\tilde{x})^{-1}\tilde{x}^TD^*} \right] \\
		=&\textup{E}\left[ \hat{\mathcal{L}}(D)-\hat{\mathcal{L}}(D^*) \right] + \textup{E} \left[\Braket{(\tilde{x}^T\tilde{x})^{-1}\tilde{x}^T(D-D^*), \epsilon} \right] \\
	\end{aligned}
\end{equation} 

In Eq. \ref{eq:conherence-proof}, the first equation arises due to the linear invariance of the gradient with respect to expectation. The second equation is derived by expanding the objective function and calculating its gradient, followed by reorganization. The third equation results from adding and subtracting the same variable $y$ on the right-hand side of the second equation. The fourth equation reconstructs the objective function and cross-terms from the third equation. The fifth equation reconstructs the objective function under optimal parameter settings from the fourth equation.

Note that the objective function under globally optimal parameter settings is necessarily less than or equal to the objective function under any other parameter settings, thus term $\textup{E}[\hat{\mathcal{L}}(D)-\hat{\mathcal{L}}(D^*)] \ge 0$. Given the previous assumption that each instance within the optimal parameter set perfectly captures the trend in $y$, the second term on the right-hand side of the fifth equation is equivalent to the previously defined zero-mean random variable $\epsilon$, and hence the second expected value is identically zero. Thus, it is proven that AOPU's optimization is coherent. 

\subsection{Proof to Proposition \ref{pro:5}}
\label{proof:pro5}

Using $dist_M$ and $dist_E$ represent Mahalanobis distance and Euclidean distance respectively we have $dist_E(x,y;\Sigma)=\sqrt{x^T\Sigma^{-1}y}$ and $dist_E(x,y)=\sqrt{x^Ty}$. The major difference between them is that the former adjusts for the distribution of data across different dimensions. Euclidean distance is essentially the Mahalanobis distance when the covariance matrix is $I$ (i.e., when dimensions are independent and identically distributed). Both Mahalanobis and Euclidean distances are strictly convex functions with respect to the input, making them suitable for use as regularizer terms in mirror maps. Referring to Algorithm \ref{alg:smd}, we have revised the training strategy for AOPU, presented in Algorithm \ref{alg:aopu-smd}. It is evident that both share a consistent optimization structure, thus structurally ensuring that AOPU's optimization process aligns with SMD. The key to the proof lies in establishing the relationship between the mirror map in SMD and the dual parameter in AOPU.
\begin{equation}
	\label{eq:smg-proof}
	\begin{aligned}
		&\nabla \left[  \Braket{\tilde{W}_n,D_n} - h(D_n) \right] \\
		=&\nabla \left[  \Braket{\tilde{W}_n,D_n} - \frac{1}{2}D_n^T(\tilde{x}\tilde{x}^T)^{-1}D_n \right]\\
		=& \tilde{W}_n-(\tilde{x}\tilde{x}^T)^{-1}D_n 
	\end{aligned}
\end{equation}

\begin{algorithm}
	\caption{SMD in AOPU}
	\label{alg:aopu-smd}
	\begin{algorithmic}[1]
		\REQUIRE Initial $D(\tilde{W})_0$
		\STATE $n \gets 0$
		\REPEAT
		\STATE $D(\tilde{W})_n = Q(\tilde{W}_n)$
		\STATE $\tilde{W}_{n+1} = \tilde{W}_n - \eta \nabla \hat{\mathcal{L}}(D(\tilde{W})_n)$
		\STATE $n \gets n + 1$
		\UNTIL{end}
		\RETURN solution candidate $\tilde{W}_n$
	\end{algorithmic}
\end{algorithm}
Clearly, the mirror map is the solution where the gradient with respect to $D_n$ is zero in $\Braket{\tilde{W}_n,D_n} - h(D_n)$. Eq. \ref{eq:smg-proof} details this gradient computation process, where the first equation is obtained by incorporating the square Mahalanobis distance into the regularizer $h$, and the second equation is derived by differentiating with respect to $D_n$. The solution $D_n=\tilde{x}\tilde{x}^T\tilde{W}$ precisely matches the definition of the dual parameter in AOPU, confirming the coherence of AOPU’s optimization strategy with the principles of SMD.

\section{Dataset Description}
\label{data descrip}
\begin{figure*}
	\centering
	\begin{adjustbox}{center}
		\includegraphics[width=1.1\columnwidth]{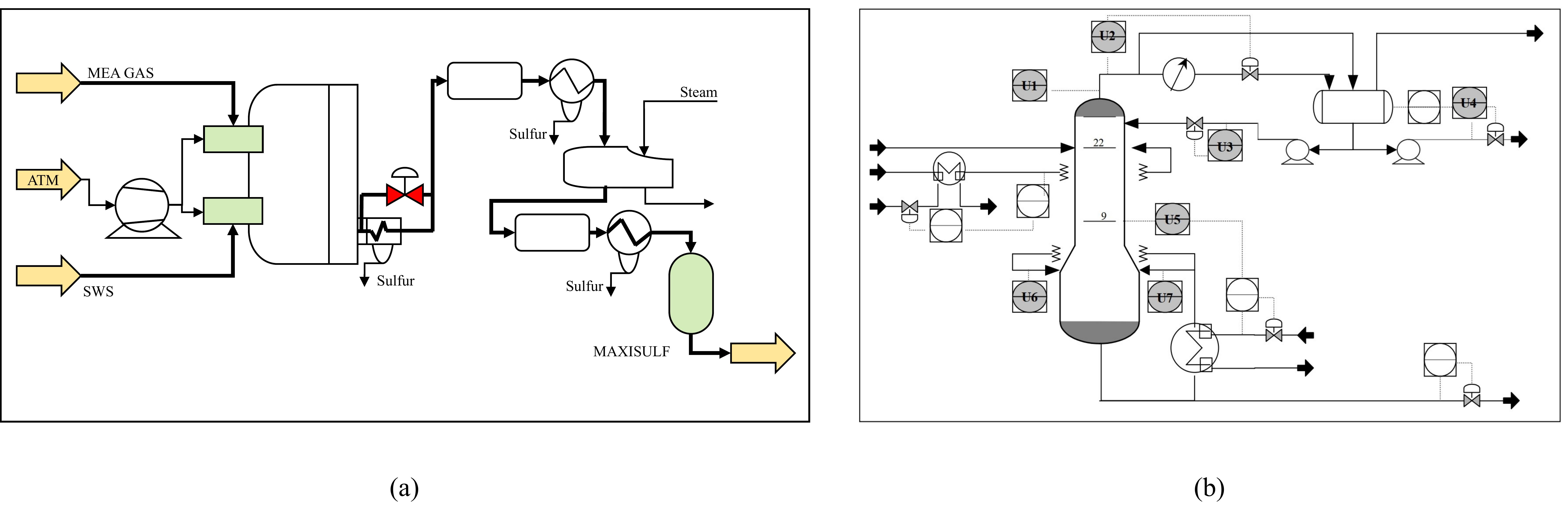}
	\end{adjustbox}
	\caption{Schematic diagram of two industrial process. (a) SRU. (b) Debutanizer.}
	\label{fig-dataset}
\end{figure*}

\begin{table*}
	\caption{Variable Description}
	\label{tab:dataset description}
	\centering
	{\scriptsize
		\setlength{\tabcolsep}{7pt}
		\begin{adjustbox}{center}
			\begin{threeparttable}
				\begin{tabular}{ccc|ccc}
					\toprule
					\multicolumn{3}{c}{Debutanizer} & \multicolumn{3}{c}{SRU} \\
					\midrule
					{Process Variables} &Unit & {Description} & {Process Variables}&Unit & {Description} \\
					U$_1$ &$^\circ \text{C}$& Top temperature & U$_1$ & $\text{m}^{3}\cdot \text{h}^{-1}$ &Gas flow MEA\_GAS \\
					U$_2$ &kg$\cdot \text{cm}^{-2}$ &Top pressure & U$_2$ & $\text{m}^{3}\cdot \text{h}^{-1}$ &Air flow AIR\_MEA \\
					U$_3$ &$\text{m}^{3}\cdot \text{h}^{-1}$ &Reflux flow & U$_3$ & $\text{m}^{3}\cdot \text{h}^{-1}$ &Secondary air flow AIR\_MEA\_2 \\
					U$_4$ &$\text{m}^{3}\cdot \text{h}^{-1}$& Flow to next process & U$_4$ & $\text{m}^{3}\cdot \text{h}^{-1}$ &Gas flow in SWS zone \\
					U$_5$ &$^\circ \text{C}$& 6$^{\text{th}}$ temperature & U$_5$ & $\text{m}^{3}\cdot \text{h}^{-1}$ &Air flow in SWS zone \\
					U$_6$ &$^\circ \text{C}$& Bottom temperature A & \quad &\quad &\quad \\
					U$_7$ &$^\circ \text{C}$& Bottom temperature B & \quad & \quad  &\quad\\
					\bottomrule
				\end{tabular}
			\end{threeparttable}
		\end{adjustbox}
	}
\end{table*}

\subsection{Debutanizer}
The Debutanizer column is part of a desulfuring and naphtha splitter plant. It is required to maximize the C5 (stabilized gasoline) content in the Debutanizer overheads(LP gas splitter feed), and minimize the C4 (butane) content in the Debutanizer bottoms (Naphtha splitter feed) \cite{Soft-sensors-for-monitoring}. However, the butane content is not directly measured on the bottom flow, but on the overheads of the downstream deisopentanizer column by the gas chromatograph resulting in a large measuring delay, which is the reason soft sensor steps in.

The dataset comprises 2,394 records, each featuring 7 relevant sensor measurements. The flowchart of the Debutanizer column, detailing the locations of these sensors and their respective descriptions, is presented in Fig. \ref{fig-dataset} (b). The corresponding details can also be found in Table \ref{tab:dataset description}.

\subsection{Sulfur Recovery Unit}
The sulfur recovery unit (SRU) removes environmental pollutants from acid gas streams before they are released into the atmosphere. The main chamber is fed with MEA gas, and combustion is regulated, in air deficiency, by supplying an adequate airflow (AIR\_MEA). The secondary combustion chamber is mainly fed with SWS gas and a suitable air flow is provided (AIR\_SWS). The combustion of SWS gas occurs in a separate chamber with excess air, in order to prevent the formation of ammonium salts in the equipment, thereby giving rise to the generation of nitrogen and nitrogen oxides. Air flows are controlled by plant operators to guarantee a correct stoichiometric ratio in the tail gas. Control is improved by a closed-loop algorithm which regulates a further airflow (AIR\_MEA\_2) on the basis of analysis of the tail gas composition. On-line analyzers are used to measure the concentration of both hydrogen sulfide and sulfur dioxide in the tail gas of each sulfur line. Hydrogen sulfide and sulfur dioxide frequently cause damage to sensors, which often have to be removed for maintenance. The design of soft sensors able to predict H2S and SO2 concentrations is therefore required.

The dataset contains 10,080 records with 5 relevant sensor measurements. The flowchart for the SRU is illustrated in Fig. \ref{fig-dataset} (a), with the input descriptions provided in Table \ref{tab:dataset description}

\section{Supplementary Figure of Training Dynamics}
\label{supp figure}

\begin{figure*}
	\centering
	\begin{adjustbox}{center}
		\includegraphics[width=.87\columnwidth]{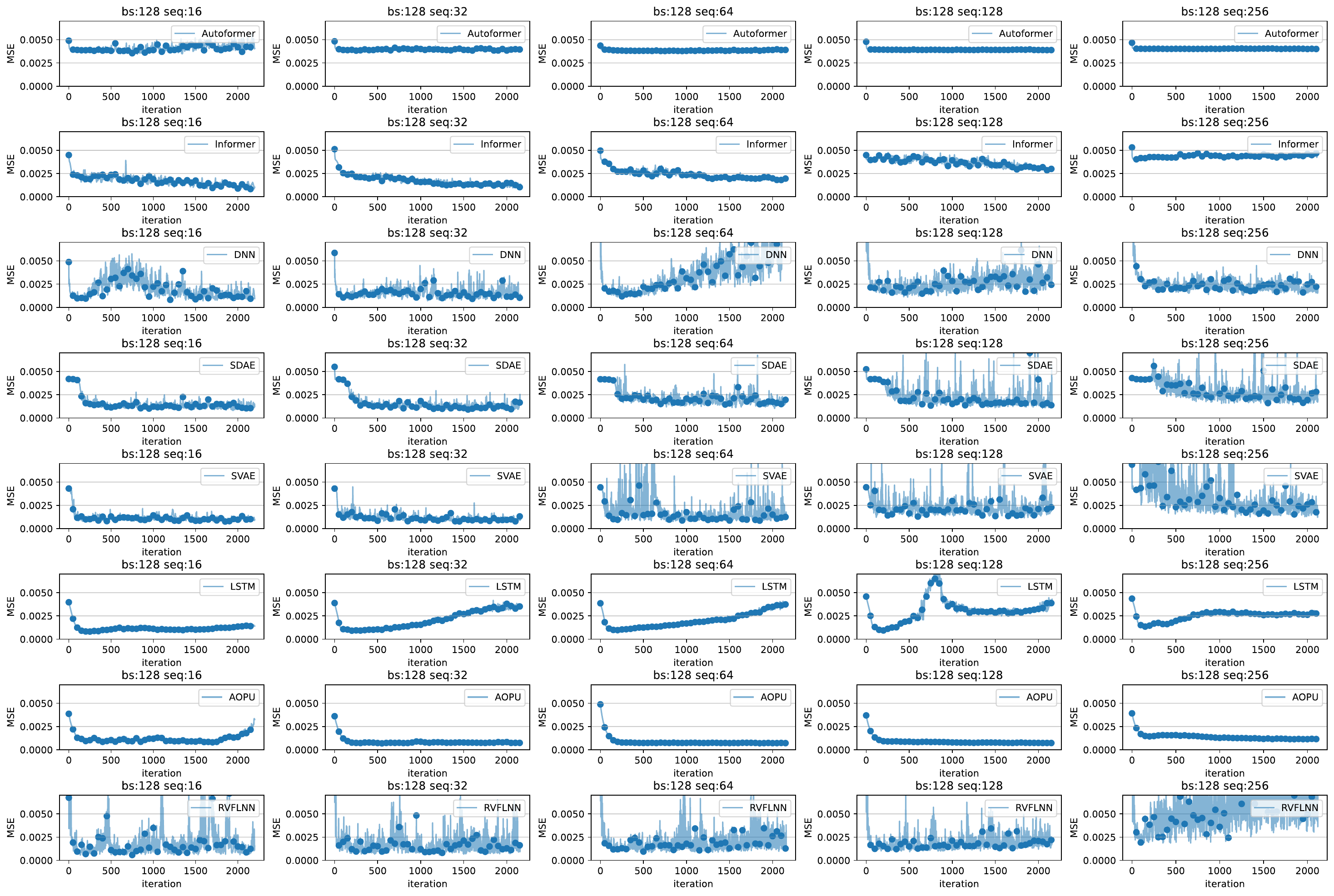}
	\end{adjustbox}
	\caption{Curves of SRU validation loss changes with training iteration for different models with fixed batch size of 128 at different sequence length settings.}
	\label{sru bs128}
\end{figure*}

\begin{figure*}
	\centering
	\begin{adjustbox}{center}
		\includegraphics[width=.87\columnwidth]{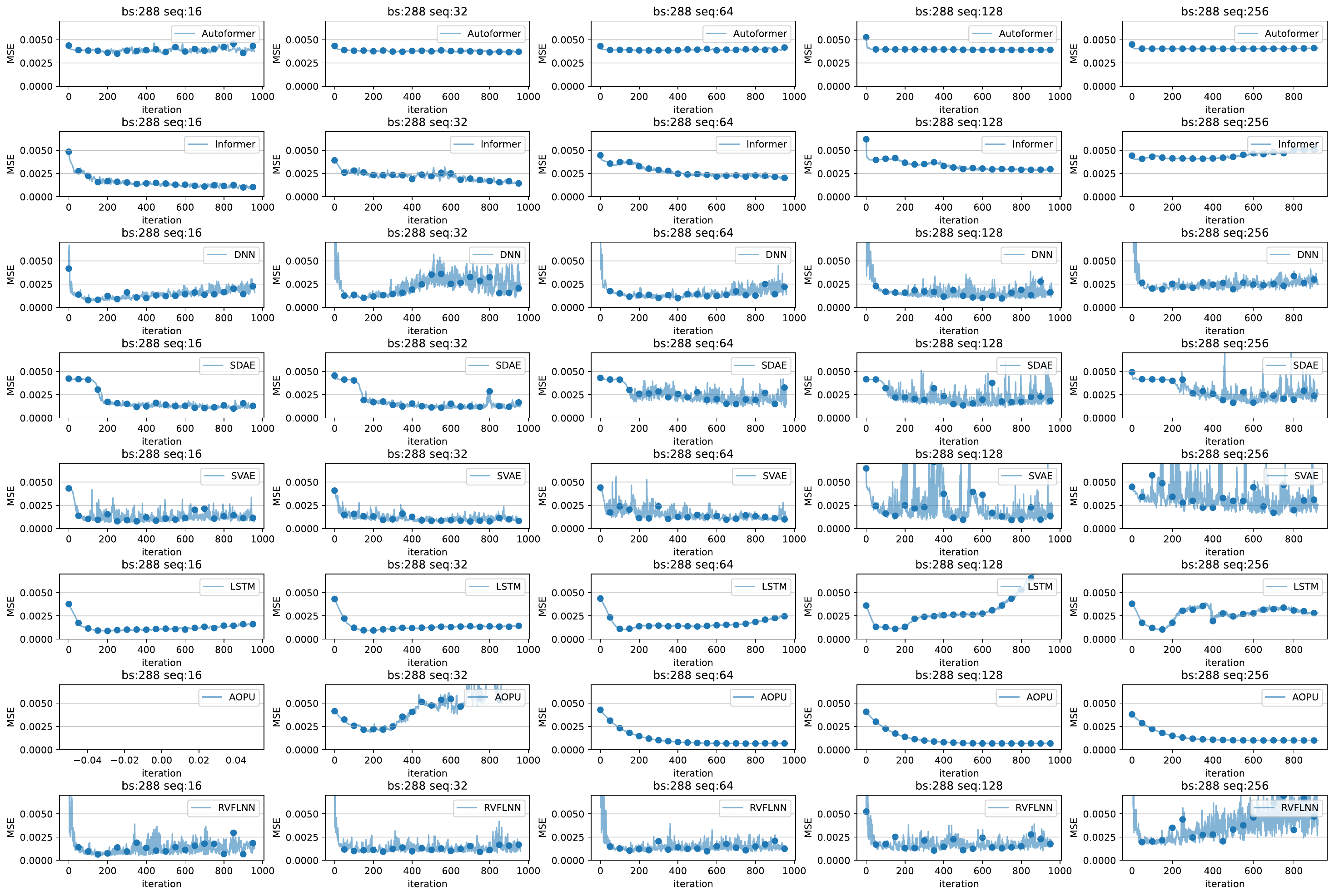}
	\end{adjustbox}
	\caption{Curves of SRU validation loss changes with training iteration for different models with fixed batch size of 288 at different sequence length settings.}
	\label{sru bs288}
\end{figure*}

\begin{figure*}
	\centering
	\begin{adjustbox}{center}
		\includegraphics[width=1\columnwidth]{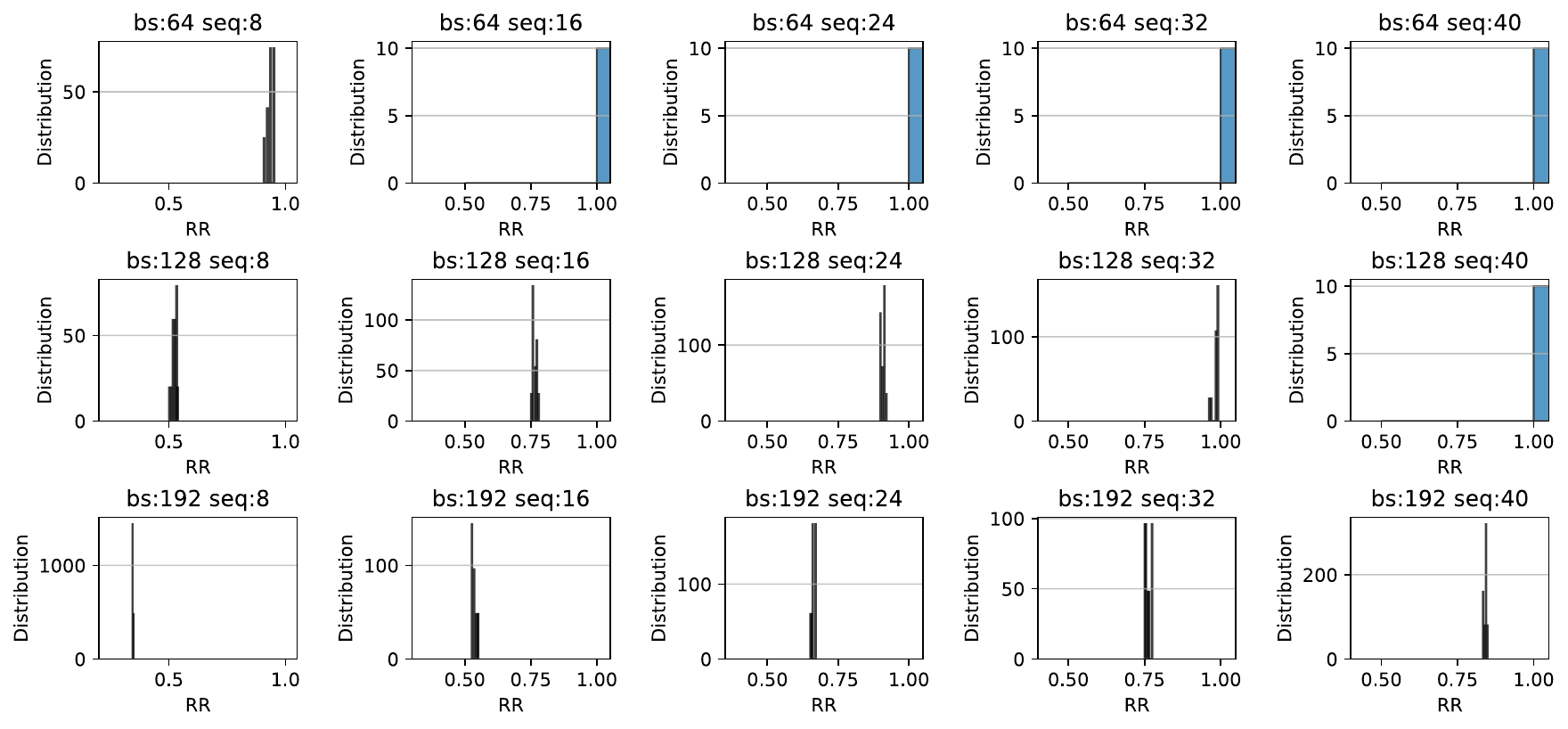}
	\end{adjustbox}
	\caption{Histogram of the frequency distribution of RR on Debutanizer dataset under varying batch sizes and sequence length settings. In each subplot, the horizontal axis represents RR, while the vertical axis indicates frequency, with the distribution normalized. Subplots within the same column have the same sequence length, while subplots within the same row have the same batch size.}
	\label{debutanizer distribution}
\end{figure*}

\begin{figure*}
	\centering
	\begin{adjustbox}{center}
		\includegraphics[width=1\columnwidth]{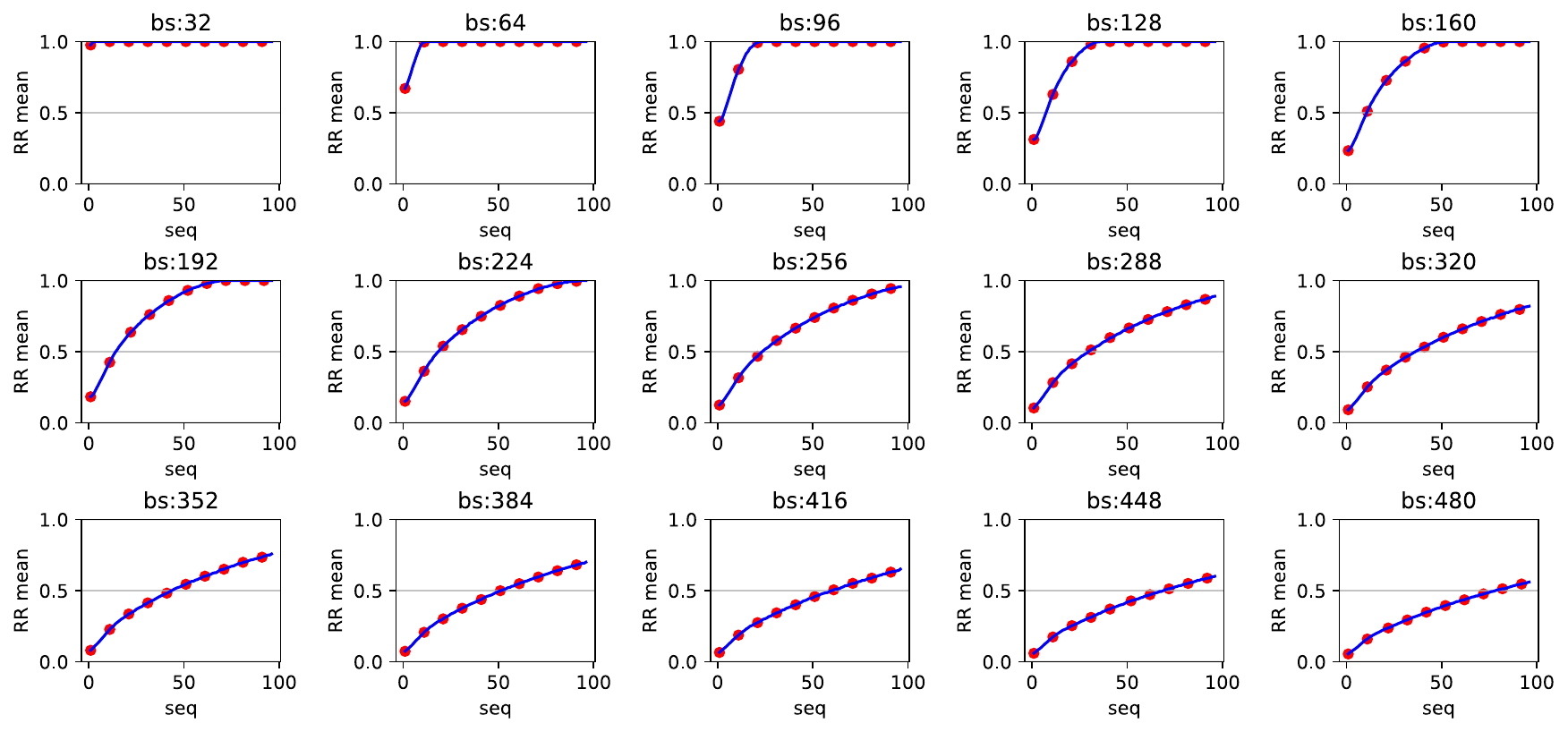}
	\end{adjustbox}
	\caption{Curve of the mean of RR distribution on Debutanizer dataset under varying batch sizes and sequence length settings. In each subplot, the horizontal axis represents sequence length, while the vertical axis indicates the mean of RR distribution. The batch size increases from left to right and from top to bottom.}
	\label{debutanizer trend}
\end{figure*}

\begin{figure*}
	\centering
	\begin{adjustbox}{center}
		\includegraphics[width=.87\columnwidth]{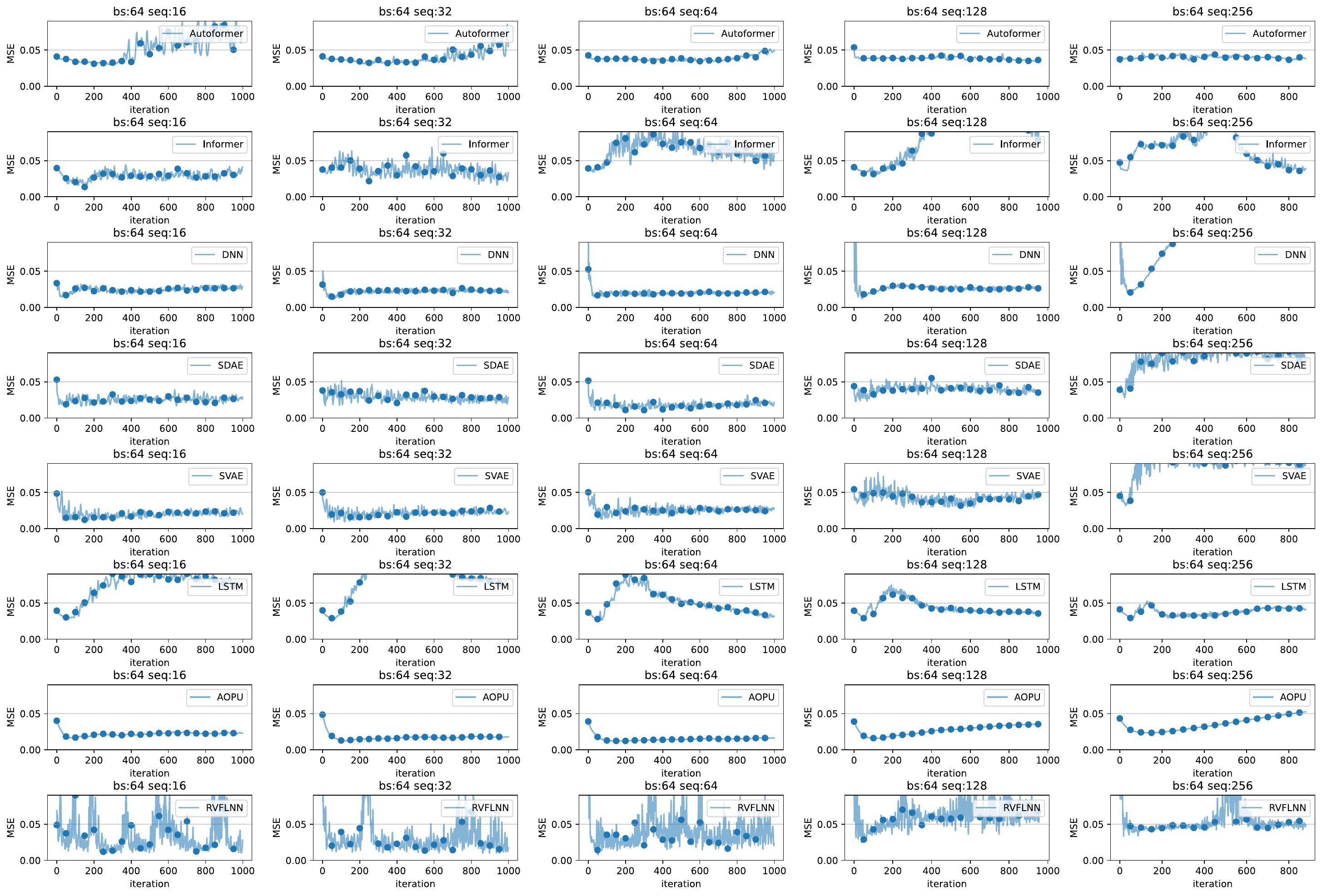}
	\end{adjustbox}
	\caption{Curves of Debutanizer validation loss changes with training iteration for different models with a fixed batch size of 64 at different sequence length settings.}
	\label{debutanizer bs64}
\end{figure*}

\begin{figure*}
	\centering
	\begin{adjustbox}{center}
		\includegraphics[width=.87\columnwidth]{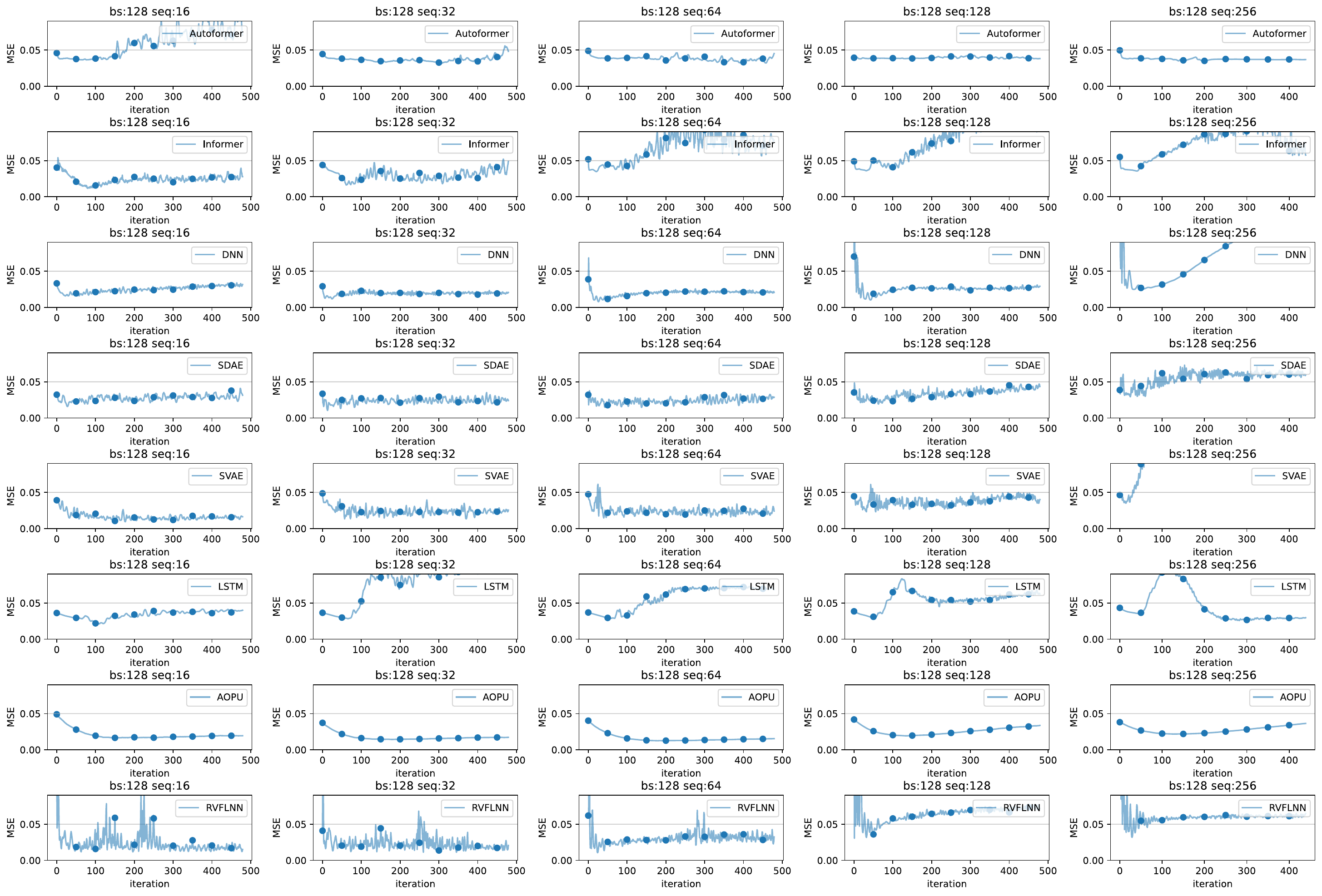}
	\end{adjustbox}
	\caption{Curves of Debutanizer validation loss changes with training iteration for different models with a fixed batch size of 128 at different sequence length settings.}
	\label{debutanizer bs128}
\end{figure*}

\begin{figure*}
	\centering
	\begin{adjustbox}{center}
		\includegraphics[width=.87\columnwidth]{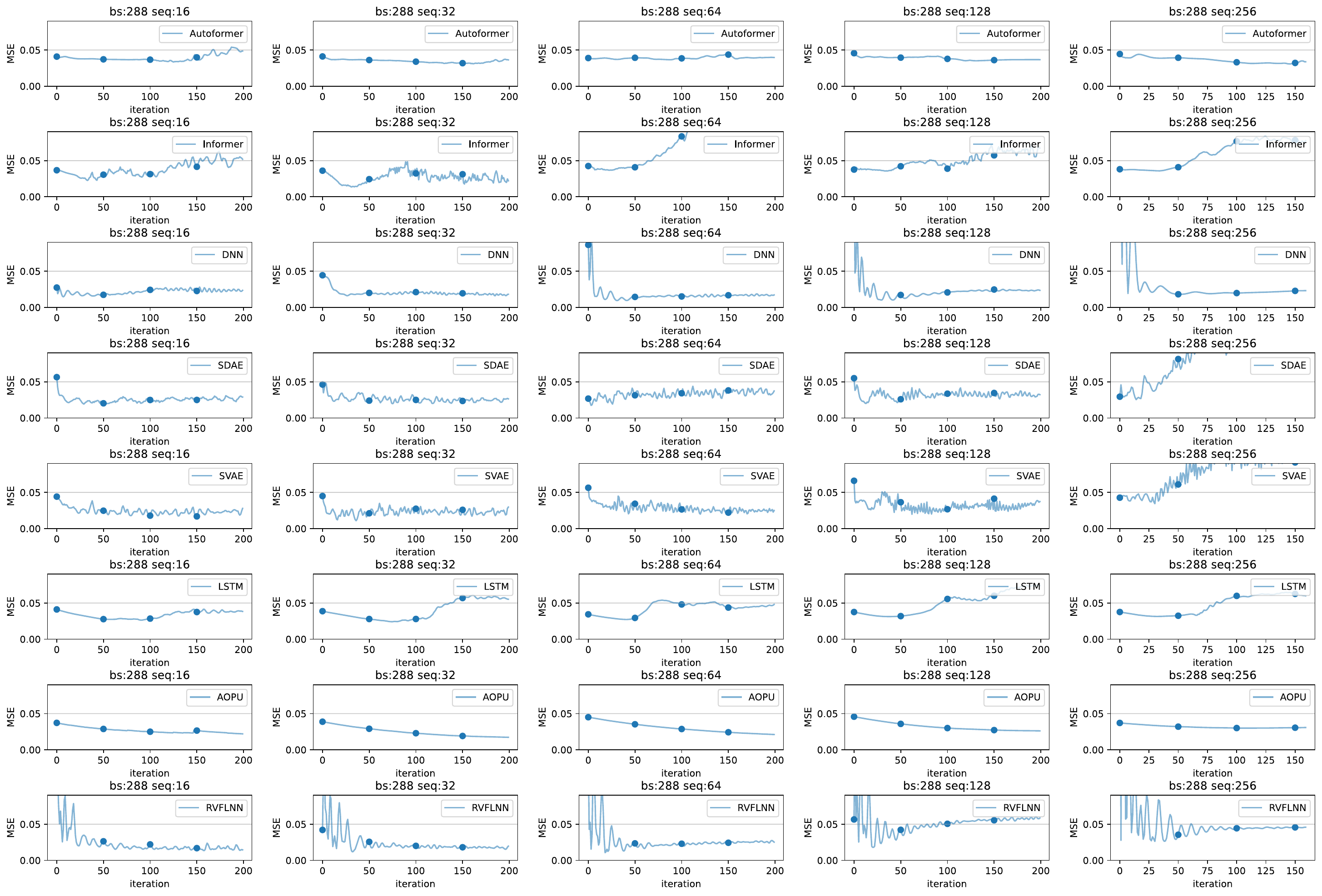}
	\end{adjustbox}
	\caption{Curves of Debutanizer validation loss changes with training iteration for different models with a fixed batch size of 288 at different sequence length settings.}
	\label{debutanizer bs288}
\end{figure*}

In this section, we complete the comprehensive RR distribution experimental results and the stability of the training process of AOPU on the SRU and Debutanizer datasets. We note that, with a fixed sequence length and increasing batch size, the variance during training generally decreases for other models, as illustrated in Fig. \ref{sru bs128} and Fig. \ref{sru bs288}. However, for AOPU, increasing bs actually diminishes performance. As seen in Fig. \ref{sru bs128}, when bs is 128 and seq is 16, AOPU's performance has declined compared to bs of 64 and seq of 16 in Fig. \ref{sru bs64}, characterized by increased fluctuations and an upward shift in the loss curve. Even more critical, Fig. \ref{sru bs288} shows that when bs is 288 and seq is 16, AOPU encounters convergence issues.

This performance degradation is due to the gradual shift of the RR distribution toward zero as bs increases. In the AOPU structure design, RR serves not only as an indicator of numerical stability during the matrix's inversion in forward computation but also as a theoretical foundation for AOPU's recovery of $y$ from $\tilde{x}y$. Therefore, lower RR correlates with poorer AOPU performance. From Fig. \ref{sru trend}, we observe that with bs at 128 and seq at 16, the mean RR is around 0.5, while at BS of 288 and sequence length of 16, the mean RR distribution is near 0.25. Given that the RR distribution is highly concentrated, the mean value represents the statistical properties of the entire distribution, indicating that AOPU cannot converge when RR falls below 0.25.

Another interpretation of AOPU's lack of convergence is that when RR is too low (e.g., assuming RR is zero), AOPU's pseudo-inverse $(\tilde{x}^T\tilde{x})^{-1}$ has been calculated as $(\tilde{x}^T\tilde{x})$, effectively converting the original normalization process $(\tilde{x}^T\tilde{x})^{-1}\tilde{x}^T\tilde{x}$ into a squared process of $(\tilde{x}^T\tilde{x})\tilde{x}^T\tilde{x}$, significantly deviating from the initial computational assumptions and causing model collapse. Similar observations can be found in the Debutanizer experiment results in Fig. \ref{debutanizer distribution}, Fig. \ref{debutanizer trend}, Fig. \ref{debutanizer bs64}, Fig. \ref{debutanizer bs128}, and Fig. \ref{debutanizer bs288}.

\section{Hyperparameter Scanning}

\begin{figure*}
	\centering
	\begin{adjustbox}{center}
		\includegraphics[width=.87\columnwidth]{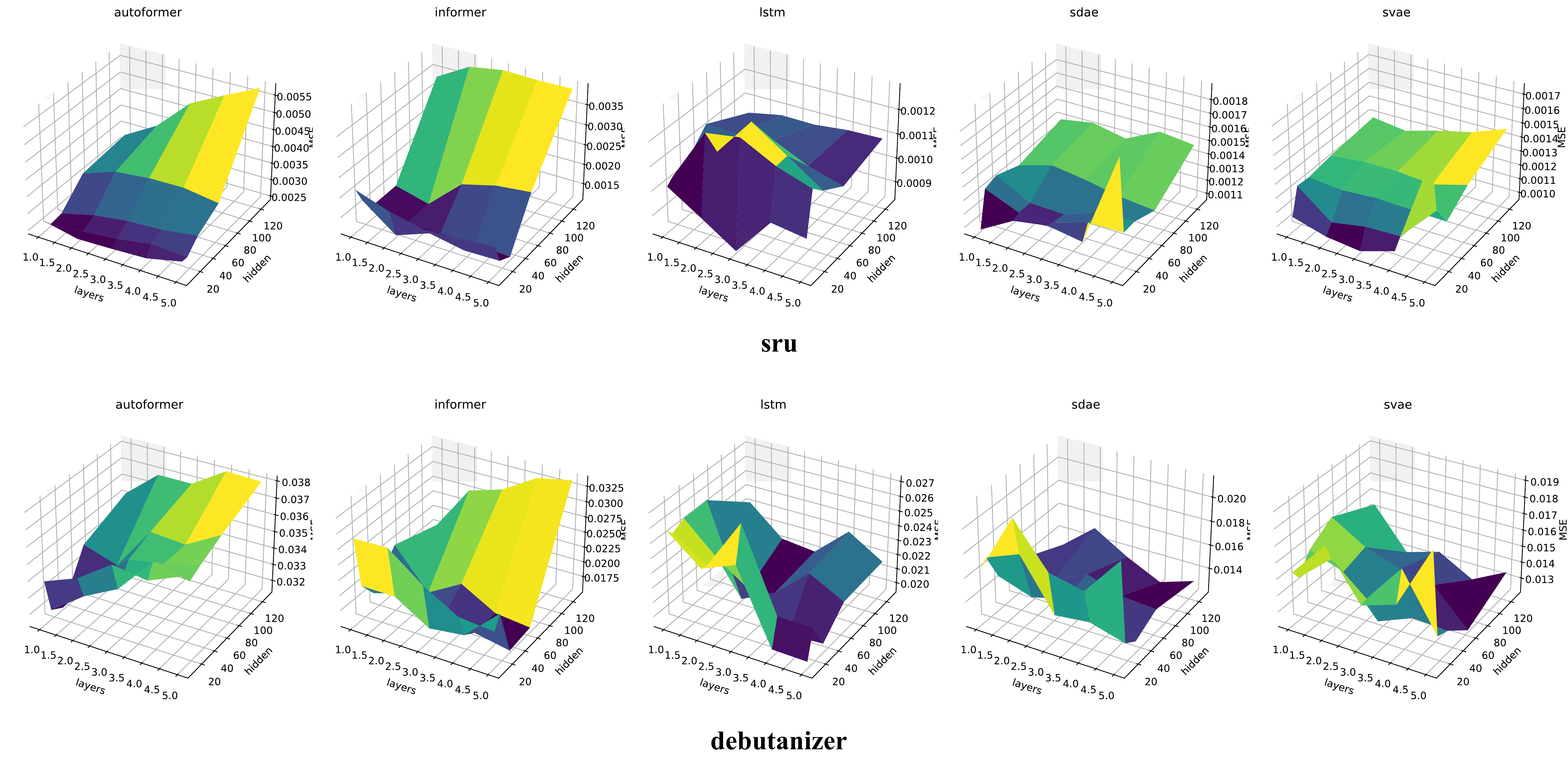}
	\end{adjustbox}
	\caption{Curves of Debutanizer validation loss changes with training iteration for different models with a fixed batch size of 288 at different sequence length settings.}
	\label{hyper tun}
\end{figure*}

\begin{figure*}
	\centering
	\begin{adjustbox}{center}
		\includegraphics[width=.87\columnwidth]{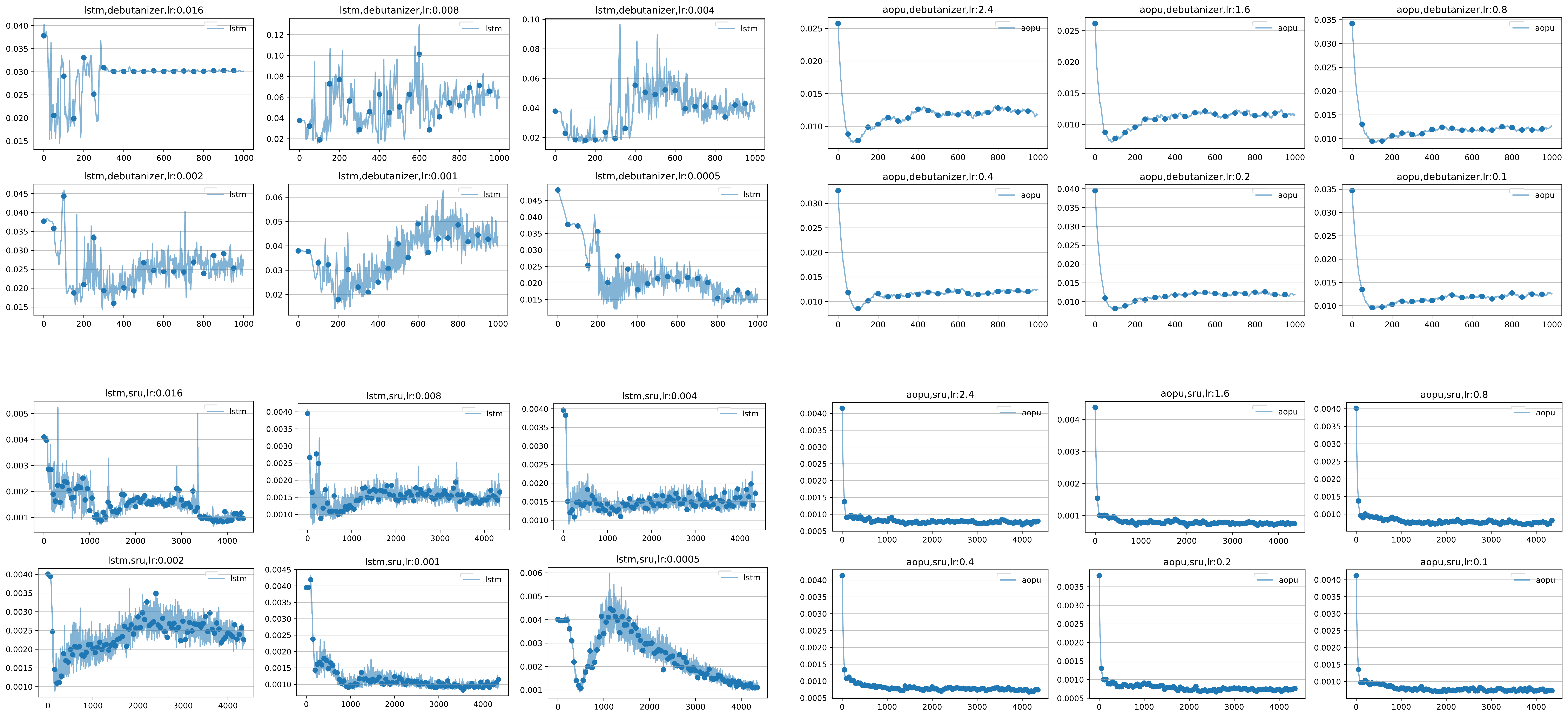}
	\end{adjustbox}
	\caption{Curves of Debutanizer validation loss changes with training iteration for different models with a fixed batch size of 288 at different sequence length settings.}
	\label{iter loss}
\end{figure*}

The hyperparameter selection is guided by two principles: first, to ensure the model size of various comparative methods remains comparable; second, to choose hyperparameters that optimize model performance. 
We present the detailed hyperparameter information in Fig. \ref{hyper tun}, where we can see for SRU the smaller setup is recommended, while for Debutanizer bigger model possibly leads to better but still limited performance. 
However, the model size of the compared methods increases dramatically with layers and hidden dims, which means that the efficiency of parameters drops. Therefore, we chose hyperparameter settings that keep the model size comparable to that of AOPU, maintaining a balance between performance and efficiency.

We paid specific attention on the learning rate (lr) setup.
To substantiate that the exceptional stability of AOPU arises from its comprehension of the manifold rather than mere fine-tuning of the learning rate, we present in Fig. \ref{iter loss} the loss trajectories over iterations under an extensive array of lr configurations. Evidently, while a reduced lr may improve model performance, it does not enhance the stability of the training process; indeed, LSTM manifests significant overfitting issues on SRU. In contrast, AOPU consistently maintains comparable stability and performance across a wide spectrum of lr variations, a merit ensured by the theoretical foundations previously expounded upon, thereby accentuating its superiority.

\end{document}